\documentclass[10pt,twocolumn,letterpaper]{article}

\usepackage[pagenumbers]{cvpr} 
\usepackage{multirow}

\definecolor{cvprblue}{rgb}{0.21,0.49,0.74}
\usepackage[pagebackref,breaklinks,colorlinks,allcolors=cvprblue]{hyperref}

\usepackage[most]{tcolorbox}

\def\paperID{30395} 
\def\confName{CVPR}
\def\confYear{2026}

\title{Cube Bench: A Benchmark for Spatial Visual Reasoning in MLLMs}

\author{
    Dhruv Anand \quad Ehsan Shareghi\\
    Department of Data Science and AI, Monash University, Victoria, Australia\\
    {\tt\small dana0001@student.monash.edu \quad ehsan.shareghi@monash.edu}
}

\begin{document}
\maketitle

\begin{abstract}

We introduce \textbf{Cube Bench}, a Rubik's-cube benchmark for evaluating spatial and sequential reasoning in multimodal large language models (MLLMs). The benchmark decomposes performance into five skills: (i) reconstructing cube faces from images and text, (ii) choosing the optimal next move, (iii) predicting the outcome of a candidate move without applying it, (iv) executing multi-step plans while recovering from mistakes, and (v) detecting and revising one’s own errors. Using a shared set of scrambled cube states, identical prompts and parsers, and a single distance-to-solved metric, we compare recent MLLMs side by side as a function of scramble depth. Across seven MLLMs, accuracy drops sharply with depth; once a trajectory stalls or diverges, models rarely recover, and high face-reconstruction accuracy does not guarantee competent action selection or multi-step execution. A pronounced closed- vs open-source gap emerges: the strongest closed model leads on both single-step perception tasks and multi-step control tasks, while open-weight models cluster near chance on the hardest settings; yet even the best MLLM degrades at higher cube complexity. A simple self-correction via reflective thinking yields modest gains but can also introduce overthinking. Cube Bench offers a compact, reproducible probe of sequential spatial reasoning in MLLMs.\footnote{The code is available at \url{https://github.com/dana-23/cube-bench}.}
\end{abstract}

\section{Introduction}
\label{section:introduction}

Multimodal large language models (MLLMs) now match or surpass prior systems on static perception benchmarks, i.e., recognizing objects, transcribing text in images, and answering short questions. However, practical deployments require models that can act interactively: taking actions, observing consequences, and staying coherent across many steps (closed-loop control). In such settings, small errors compound and recovery is rare; single-shot vision tests provide little signal about whether a model can plan, act, and correct itself. Surveys of multimodal benchmarks highlight this mismatch between static leaderboards and the demands of interactive use \cite{li2024surveybenchmarksmultimodallarge}, and recent web-based agent evaluations similarly report that strong snapshot performance does not directly translate to reliable multi-step behaviour \cite{zhou2024webarenarealisticwebenvironment,koh2024visualwebarenaevaluatingmultimodalagents}. Classical work on sequential decision making has long emphasized how compounding errors degrade long-horizon performance \cite{Goyal2017Making}.

\begin{figure}[t]
    \centering
    \includegraphics[trim={0.55cm 0.25cm 0.55cm 0.3cm},clip, width=\columnwidth]{Images/Cube_Bench.pdf}
    \caption{\textbf{Cube Bench} flow: Given depth and seed, a \emph{VirtualCube} state yields (i) a rendered image, (ii) a canonical text state, and (iii) a four-option set (A–D). The MLLM consumes the modalities and prompt, and returns output: \texttt{Yes|No} (Verification), \texttt{A|B|C|D} (Move Prediction / Closed-Loop), or \texttt{DECREASE|NO\_CHANGE|INCREASE} (Causal Move-Effect).}
    \label{fig:cubebench}
\end{figure}

Existing visual and spatial reasoning benchmarks mostly operate in this static regime. Synthetic datasets such as CLEVR, GQA, and related visual question answering benchmarks probe fine-grained spatial relations, counting, and logical composition from a single image, but typically stop at one-shot answers about a fixed scene. More recent multimodal leaderboards continue this pattern, emphasizing captioning, OCR, and short-form VQA on web images \cite{li2024surveybenchmarksmultimodallarge}. These settings test local spatial understanding, yet they rarely couple perception to an explicit dynamics model, track progress toward a goal, or require models to maintain a consistent internal state over multiple actions. On the other side, rich agentic environments on the web or in 3D scenes \cite{zhou2024webarenarealisticwebenvironment,koh2024visualwebarenaevaluatingmultimodalagents} introduce realistic noise and partial observability, but success metrics are often coarse, and it is difficult to disentangle failures of perception, planning, and control. We therefore lack a compact, generator-based domain that is simultaneously spatial and sequential, fully observed, exactly solvable, and regenerable on demand with strict fairness controls and oracle progress signals, making it possible to localize failures within a decision loop rather than aggregate them into a single static score.

We introduce \textbf{Cube Bench}, a compact and fully controllable generator-based benchmark for studying long-horizon decision making under complete state information (see Figure~\ref{fig:cubebench}). Conceptually, Cube Bench instantiates a simple \emph{TEA} loop i.e., \emph{Transition} $\rightarrow$ \emph{Evaluation} $\rightarrow$ \emph{Argmin} (Action): the cube dynamics define an exact transition model, oracle distance provides a scalar evaluation of each state, and the model’s role is to choose actions that (ideally) minimize distance to the goal at each step. Cube Bench uses the Rubik's Cube as a testbed: it is visually simple, easy to simulate exactly, and combinatorially rich, with known optimal and near-optimal solvers that tell us precisely how many moves remain to reach the goal \cite{korf1997finding,rokicki2014diameter}. This lets us, unlike static image corpora, generate unlimited episodes, render both an image and a concise text description of each state, present a small menu of candidate moves, and compute exact feedback on how much each move improves or worsens the situation.

On top of this testbed we define seven tests that target different stages of the \emph{see $\rightarrow$ evaluate $\rightarrow$ act $\rightarrow$ reflect $\rightarrow$ recover} loop: (i) reconstructing the full state from vision; (ii) checking consistency between vision and text; (iii) choosing the next move from a short list; (iv) judging whether a move helps or harms progress; (v) executing a sequence of moves to reach the goal; (vi) using self-reflection to revise an earlier choice; and (vii) trying to recover after the first error. Our framework enforces strict fairness controls, yielding measurements that are easy to replicate, difficult to game, and tailored to disentangle which stage of the TEA loop is failing.

Evaluating strong open- and closed-source MLLMs under identical rules reveals a consistent picture. First, \emph{seeing is not reasoning}: models that parse cube faces well still struggle to sustain improvement over multiple steps. Second, \emph{feedback breaks coherence}: when outputs are fed back as inputs, error rates rise sharply with scramble depth, and recovery after a wrong move is rare. Third, \emph{reflection helps but is fragile}: structured, label-aware reflection can repair some mistakes, yet unconstrained reflection can induce overthinking and instability \cite{shinn2023reflexion,madaan2023selfrefine,yao2023react}. Together, these results pinpoint lingering weaknesses in sequential spatial reasoning, i.e., state tracking and action evaluation, that are not revealed by current single-step benchmarks or coarse web-agent success rates.

\section{Related Work}
\label{section:related_work}
\noindent\textbf{Static perception benchmarks for MLLMs.}
A large body of work evaluates multimodal LLMs on single-shot perception tasks such as classification, detection, captioning, VQA, OCR, and charts, e.g., ImageNet, COCO, CLEVR, GQA, TextCaps, and ChartQA \cite{Russakovsky2015ImageNet,Lin2014MicrosoftCOCO,Johnson2017CLEVR,Hudson2019GQA,Sidorov2020TextCaps,Masry2022ChartQA}. Recent surveys synthesize these trends and note rapid leaderboard gains driven by scaling and instruction tuning \cite{li2024surveybenchmarksmultimodallarge}. Classic perception datasets (e.g., VQAv2) emphasize recognition under short contexts and have pushed models toward improved visual grounding and bias control \cite{Goyal2017Making}; ScienceQA further probes multimodal reasoning with thought chains \cite{lu2022learn}. However, these settings rarely require \emph{closed-loop} decision making, where actions alter the state and errors can compound across steps. Our benchmark targets this gap by measuring whether models can sustain improvements over multiple actions, given precise state feedback.

\noindent\textbf{Interactive / agent benchmarks.}
Web-scale agent environments assess language-to-action competence in long-horizon tasks (navigation, form-filling, retrieval, tool use) \cite{zhou2024webarenarealisticwebenvironment,koh2024visualwebarenaevaluatingmultimodalagents}. While such benchmarks capture realistic challenges (dynamic pages, delayed rewards, partial observability), they introduce confounds, non-deterministic layouts, fragile tool chains, and evaluation noise, that complicate causal analysis of failures. We instead adopt a compact, fully controllable domain with deterministic transitions and exact progress signals, allowing us to isolate where breakdowns arise (perception, evaluation, selection, or reflection) without web-induced variance.

\noindent\textbf{Reasoning and control in structured domains.}
Structured puzzles and games offer precise state representations, small action alphabets, and well-defined optimality notions. For the Rubik’s Cube, shortest-path structure and the diameter (“God’s number”) enable oracle feedback about optimal move distance \cite{korf1997finding,rokicki2014diameter}. Learning-to-search systems such as DeepCubeA demonstrate scalable RL+search on the cube and related combinatorial puzzles \cite{agostinelli2019deepcubea}. We repurpose these properties to build an \emph{evaluation}, not a solver, so perception and control can be disentangled and measured reproducibly.

\noindent\textbf{Rubik’s-cube with VLM/LLM and embodied control.}
Recent cube-focused VLM/LLM efforts mix perception, planning, and tooling (e.g., CubeRobot, VisionCube, Auto-RubikAI), while earlier robotics work (Dactyl) highlights actuation confounds \cite{wang2025cuberobot,wang2025visioncube,fan2025autorubikai,openai2019solvingrubikscuberobot}. Our setting keeps the environment compact and oracle-scored to avoid attribution ambiguity.

\noindent\textbf{Self-reflection and tool-augmented prompting.}
Interleaving reasoning and acting (or adding self-critique) can improve sample efficiency but may also induce instability or “overthinking.” Representative approaches include Reflexion \cite{shinn2023reflexion}, Self-Refine \cite{madaan2023selfrefine}, ReAct \cite{yao2023react}, and analyses of reflection effects \cite{renze2024self}. Our experiments use a constrained, label-aware reflection protocol to quantify when reflection repairs errors and when it degrades coherence, under identical rules across models.

\noindent\textbf{Positioning.}
Compared to perception-centric benchmarks \cite{Russakovsky2015ImageNet,Lin2014MicrosoftCOCO,Johnson2017CLEVR,Hudson2019GQA,Sidorov2020TextCaps,Masry2022ChartQA,li2024surveybenchmarksmultimodallarge,Goyal2017Making} and open-web agent suites \cite{zhou2024webarenarealisticwebenvironment,koh2024visualwebarenaevaluatingmultimodalagents}, our benchmark provides (i) deterministic generation with paired image and text state, (ii) strict I/O with a small, discrete action set, (iii) fairness controls that remove answer position bias, and (iv) exact oracle progress signals grounded in optimal-move distance \cite{korf1997finding,rokicki2014diameter}. This design yields compact, reproducible measurements of closed-loop competence and exposes where modern MLLMs fail to maintain multi-step coherence, even when their single-step perception appears strong.

\section{Cube Bench}
\label{section:cubebench}
Cube Bench is a compact, fully controllable benchmark for long-horizon decision making, using the Rubik's Cube as a testbed. The cube gives us a visually simple but combinatorially rich environment with an exact optimal solver: for any state we know how many moves remain to reach the goal. This lets us study the full \emph{see $\rightarrow$ evaluate $\rightarrow$ act $\rightarrow$ reflect $\rightarrow$ recover} loop under complete state information, without web-scale noise or ambiguous evaluation metrics.

Cube Bench is simulator-based and is generated on-the-fly rather than shipping a fixed dataset. This keeps the benchmark lightweight to distribute and extend while still making results exactly reproducible across models. For a target distance $d$ (number of optimal moves from solved) and episode index $i$, the \texttt{VirtualCube} simulator deterministically emits: (i) a rendered image, (ii) an authoritative textual state, and (iii) four candidate actions (A–D). We publish per-depth seed lists; episode $i$ uses \texttt{seed}=$i$, fixing the scramble $s_{d,i}$, the teacher inverse plan $\pi^{\text{teach}}$, and the oracle distances shared across models. Oracle distance $d(s)$ is computed once via IDA* with pattern databases \cite{korf1997finding}; unless stated otherwise, we use the standard face-turn / half-turn move metric (HTM/FTM), and we use the same metric both scrambling and evaluation. To avoid answer-position bias, we enforce that the correct option is approximately uniform over A-D at each depth; the harness tracks the per-slot counts, discards duplicate items, and regenerates episodes until the distribution is near-uniform. Model outputs must follow strict one-line formats; anything else is treated as an error. Cube Bench is test-only: we release prompts, seed lists, and quality-control thresholds/methods in the \emph{Supplementary Material} for reproducibility.
\subsection{Tasks}
\label{subsection:methodology:tasks}
We evaluate seven tests that instantiate the \emph{see $\rightarrow$ evaluate $\rightarrow$ act $\rightarrow$ reflect $\rightarrow$ recover} steps.
Each episode is constructed from a published seed list at depth $d$ and exposes (i) an explicit text state (canonical cube notation), (ii) a paired image for realism, and (iii) four candidate actions $A$–$D$ (moves), with exactly one correct closed-form label per test instance.
Text is the source of truth; images never override the textual state.
We use a single move metric $d(\cdot)$ (e.g., half-turn or face-turn metric) consistently for scrambling and scoring. Next, we introduce the evaluated tasks:

\vspace{-0.75em}
\paragraph{Cube Face Reconstruction.} \emph{This test measures raw visual parsing of the cube without help from text or action selection.}
From a seed at depth $d\in\{1,2,3\}$, \texttt{VirtualCube} renders a fixed-pose cube-net image (see Supplementary for an example image); the center of the front face is labeled \texttt{Front}. The model must return a strict $3\times3$ grid for the \emph{front face only}, using color initials in \{W,Y,R,O,G,B\} row-by-row; no textual state is provided and free-form prose is disallowed.
Primary metrics are \textbf{Element-wise Acc.} (per-sticker on the front face) and \textbf{Matrix Acc.} (share of items with the exact $3\times3$ grid) reported per depth with Wilson 95\% CIs.

\vspace{-0.75em}
\paragraph{Cross-Modal Verification (image $\leftrightarrow$ text consistency).} \emph{This test measures whether vision and text agree on the same face without granting credit for format violations or label bias.} Each item pairs a fixed-pose cube-net image with a textual $3\times3$ front-face grid; the model outputs \texttt{Answer: Yes} if every cell matches and \texttt{Answer: No} otherwise. Positives are true (image, text) pairs; negatives are formed by applying $1\text{–}3$ front-face token edits to the text or by rolling out exactly one oracle move on the text while keeping the image fixed. All items are drawn from cube states at oracle distance $d\!=\!5$ from solved, and we sample an equal number of \verb|Yes| and \verb|No| cases.
Primary metrics are (1) \textbf{Bal} ($=(\mathrm{TPR}+\mathrm{TNR})/2$) where TPR and TNR denote True Positive and Negative Rates, (2) \textbf{Parse} which is the percentage of outputs matching \texttt{Answer: Yes|No}, and \textbf{Bias} ( $=\lvert\text{\texttt{Yes} rate}-0.5\rvert$) where lower value indicate less systematic bias toward answering \texttt{Yes}.

\vspace{-0.75em}
\paragraph{Optimal Move Prediction.} \emph{This test measures one-step action choice at the point of execution, disentangled by prompt modality.}
From a canonical textual state at $d\!=\!1$, the model is given four candidate moves $A\text{–}D$ and must select the single best option in \{A,B,C,D\}. At $d\!=\!1$ exactly one face turn is the inverse scramble; we sample three legal but suboptimal distractors and shuffle answer slots to keep the correct option roughly uniform over A-D. The identical item is presented under three prompt modalities: Image+Text, Image-only, and Text-only. No rollout occurs and strict parsing requires either \texttt{[A–D]} or \texttt{<ANSWER>…</ANSWER>}. 
Primary metrics are \textbf{Acc.} (Top-1) and \textbf{Parse} (percentage of outputs in the allowed format), which are reported per modality with matched item counts.

\vspace{-0.75em}
\paragraph{Reflection-Guided Re-Answering.}\emph{This test measures whether a structured self-check can repair wrong answers without destabilizing correct ones.}
Using the same Optimal Move Prediction items, we take the model’s initial choice and run a single reflection pass under two regimes: Guided (Unredacted), which reveals the correct option during reflection to estimate an oracle upper bound, and Guided (Redacted), which hides labels (default). Prompts, parsing, and item balancing mirror Optimal Move Prediction; decoding uses zero temperature, deterministic option shuffling, and strict $A\text{–}D$ parsing.
Primary metrics reported for both Unredacted and Redacted settings are \textbf{Initial Acc.}, \textbf{Final Acc.}, $\mathbf{\Delta}=\text{Final}-\text{Initial}$, \textbf{EFR} (percentage of answers initially incorrect but corrected after reflection) 
, and \textbf{OTR} (percentage of answers initially correct but flipped to incorrect after reflection). 

\vspace{-0.75em}
\paragraph{Closed-Loop Step-by-Step (act under feedback).} \emph{This test measures whether a model can \emph{maintain} a monotonic path when its outputs feed back into inputs.}
An episode starts from $s_1$ at optimal-move distance $d$ from solved.
At each step $t\in\{1,\dots,d\}$ the model selects one action from \{A–D\}.
We execute the chosen move, obtain $s_{t+1}$, and recompute $d(s_{t+1})$.
Progress is recorded when $a_t$ belongs to the optimal action set $A^\star_t=\{a:\,d(s_{t+1}) \text{ is minimal}\}$.
The rollout stops at the first non-progress step (no $a_t\in A^\star_t$) or on any parse/format violation; solved episodes terminate immediately on reaching distance $0$.
Primary metrics are \textbf{TA \%} ($=\frac{1}{d}\sum_{t=1}^{d}\mathbf{1}\{a_t\in A^\star_t\}$) and \textbf{Perfect-Solve \%} (share of episodes solved before termination). 

\begin{table*}[t]
\footnotesize
    \centering
    \resizebox{0.7\textwidth}{!}{
    \begin{tabular}{@{}l *{3}{c} *{3}{c} *{3}{c}@{}}
        \toprule
        & \multicolumn{3}{c}{\textbf{Element-wise Acc. }}
        & \multicolumn{3}{c}{\textbf{Matrix Acc. }}
        & \multicolumn{3}{c}{\textbf{Verification (\%) @ $d=5$}} \\
        \cmidrule(lr){2-4}\cmidrule(lr){5-7}\cmidrule(lr){8-10}
        \textbf{Model}
        & $d=1$ & $d=2$ & $d=3$
        & $d=1$ & $d=2$ & $d=3$
        & Bal. & Parse & Bias \\
        \midrule
        Gemma 3                 & 61.9    & 47.2   & 36.60        & 31.00   & 8.00   & 3.00       & 58.00  & 100.00 & 70.00    \\
        GLM-4.5V                & 100.00  & 99.70  & 98.70        & 100.00  & 98.00  & 96.00      & 95.00  & 100.00 & 55.00    \\
        Llama4 Scout-17B        & 70.60   & 41.00  & 33.20        & 36.00   & 7.00   & 3.00       & 47.00  & 100.00 & 93.00    \\
        Qwen2.5-VL 7B           & 76.20   & 70.60  & 62.00        & 40.00   & 22.00  & 13.00      & 71.30  & 98.00   & 57.10   \\
        Qwen2.5-VL 32B          & 84.00   & 78.80  & 71.30        & 68.00   & 43.00  & 19.00      & 81.00  & 100.00  & 65.00   \\
        Qwen3-Thinking 30B      & 98.10   & 95.40  & 86.70        & 94.00   & 78.00  & 60.00      & 84.00  & 100.00  & 66.00   \\
        Gemini 2.5 Pro          & 100.00  & 98.80  & 96.80        & 100.00  & 90.00  & 84.00      & 100.00 & 100.00 & 52.00    \\
        \bottomrule
    \end{tabular}}
    \caption{Left: \emph{Cube Face Reconstruction} reported as element-wise (per-sticker) accuracy and exact 3×3 matrix accuracy at depths $d\in\{1,2,3\}$. Right: \emph{Cross-modal Yes/No Verification} at fixed depth $d=5$ with \textbf{Bal}=$( \mathrm{TPR}+\mathrm{TNR})/2$, \textbf{Parse}=\% outputs matching \texttt{Answer: Yes|No}, and \textbf{Bias}=$|\mathrm{\texttt{Yes}\ rate}-0.5|$ (lower is better).}
    \label{tab:combined-recon-verif}
\end{table*}

\vspace{-0.75em}
\paragraph{Causal Move-Effect (pre-action evaluation).}\emph{This test isolates the model’s ability to evaluate action consequences independent of action selection.}
The MLLM is given the current state and a specific candidate move to predict the causal label in $\{\textsc{decrease},\textsc{no\_change},\textsc{increase}\}$ for the change in optimal distance $d(\cdot)$ \emph{prior} to move execution.
Labels are computed by applying the candidate move in the simulator and comparing $d(s_{t+1})$ to $d(s_t)$ under the same metric used for scrambling.
Primary metrics are \textbf{Micro-Accuracy}, \textbf{Macro-F1} over the 3 classes, and Cohen’s $\mathbf{\kappa}$ (chance-corrected agreement given empirical marginals).

\vspace{-0.75em}
\paragraph{Learning-Curve / Recovery (post-error).}\emph{This test measures whether a model can \emph{repair} trajectories once it has deviated from optimality.}
Starting from the \emph{first} Closed-Loop failure point (first non-progress step or parse error), we continue from the resulting post-error state for at most \texttt{max\_attempts} additional decisions.
At each attempt the model again chooses from \{A–D\}; we always execute its choice (even if non-optimal) to test whether it can re-enter a monotonic regime and eventually solve the puzzle.
The episode ends on solve or when the attempt budget is exhausted.
Primary metrics are
\textbf{Solve Rate (SR)} with Wilson 95\% CI; early recovery $\mathbf{P(1)}$ and $\mathbf{P(\le3)}$ (probability of solving within 1 or 3 attempts); Med@Solved (median attempts among solved episodes); Avg@All (mean attempts over all episodes, counting unsolved as \texttt{max\_attempts}). Exact $d$-dependent horizons and \texttt{max\_attempts} are provided in the Supplement.

\subsection{Episode Generation}
\label{subsection:methodology:generation_qc}
Episodes and general quality control (QC) follow what stated earlier in this section. 
Task-specific knobs are:
(i) \textbf{Closed-Loop} uses the same turn metric as scrambling, terminates at the first non-progress or parse failure (unconditional denominators), and enforces one-token A-D outputs.
(ii) \textbf{Move-Effect} uses the label set \{DECREASE, NO\_CHANGE, INCREASE\}.
(iii) \textbf{Learning-Curve} starts from the post-error state and grants \texttt{max\_attempts}.
Baselines use deterministic decoding (temperature$=$0); any deviations are declared in §\ref{section:experiments}.

\section{Experiments}
\label{section:experiments}
\paragraph{Models \& Inference Setup.}
We evaluate a range of open-source MLLMs, including Gemma 3-27B~\cite{team2025gemma}, GLM-4.5V~\cite{vteam2025glm45vglm41vthinkingversatilemultimodal}, Llama 4-Scout-17B:16E~\cite{meta_llama4_scout_17b_16e_2025}, Qwen2.5-VL-7B/32B~\cite{bai2025qwen2}, Qwen3-VL-30B-A3B~\cite{qwen3technicalreport}, and one API model, Gemini 2.5 Pro~\cite{google_gemini_2_5_pro_2025}. For engine and decoding, we utilize vLLM~\cite{kwon2023efficientmemorymanagementlarge} with bf16 and a temperature of 0 (i.e., greedy decoding) unless specified otherwise, with reflection variants specifying their own decoding. Request-level batching is used to optimize throughput. Our hardware setup consists of NVIDIA A100-80GB GPUs. To ensure reproducibility, we use identical prompts for each task and modality, publish seed lists for each depth, and maintain a fixed turn metric (HTM/FTM) per run. Parsing is strict, requiring one-line outputs formatted as \texttt{A-D}, \texttt{Yes|No}, or \texttt{|DECREASE|NO\_CHANGE|INCREASE|}, with any invalid lines counted as errors. For sampling, the per-condition \(N\) is fixed, and we reuse the same episode set across models to facilitate paired comparisons.

\subsection{Cube Face Reconstruction and Cross-Modal Verification}
\label{subsec:exp:reconstruction_verfication}
In Table~\ref{tab:combined-recon-verif} (left), by $d{=}3$ Gemini~2.5~Pro retains high per-sticker fidelity (96.8\%) with a moderate drop in exact $3\times3$ matches (84.0\%), whereas Qwen~2.5-VL-32B declines more sharply (71.3\% / 19.0\%), indicating weaker global binding. The gap between element-wise and matrix accuracy widens with depth across models, consistent with small edge misplacements compounding into full-face mismatches. Complementing this, Cross-Modal Verification at fixed $d{=}5$ (Table~\ref{tab:combined-recon-verif}, right) shows that models with near-perfect parsing still differ markedly in discrimination: Qwen3 reaches Bal.\ 84.00\%, Qwen~2.5-VL-32B 81.00\%, Qwen~2.5-VL-7B 71.30\%, GLM 95.00\%, and Gemini 100.00\%. Despite Parse rates $\approx$100\% for the top models, residual label bias persists (52-93\%); for instance, Llama~4-Scout-17B:16E shows 93\% bias, suggesting a tendency to default to \texttt{Yes} when uncertain.

\noindent\textbf{Takeaways:}
Face Reconstruction establishes the \emph{vision floor}: depth stresses spatial coherence beyond token recognition, and exact $3\times3$ matches are the first to fail. Verification then serves as a high-precision gate on cross-modal grounding, revealing discrimination gaps and label-selection biases even when parsing is flawless. We, therefore, place Verification before move selection or closed-loop tests and condition downstream analyzes, Move Prediction and Closed-Loop Recovery, on passing Reconstruction, while reporting both element-wise and matrix accuracy to separate recognition from spatial arrangement.
\begin{table}[t]
    \centering
    \resizebox{\columnwidth}{!}{%
    \begin{tabular}{@{} l *{2}{c} *{2}{c} *{2}{c} @{}}
        \toprule
        & \multicolumn{2}{c}{\textbf{Image + Text}}
        & \multicolumn{2}{c}{\textbf{Image}}
        & \multicolumn{2}{c}{\textbf{Text}} \\
        \cmidrule(lr){2-3}\cmidrule(lr){4-5}\cmidrule(lr){6-7}
        \textbf{Model}
        & \textbf{Acc. } & \textbf{Parse}
        & \textbf{Acc. } & \textbf{Parse}
        & \textbf{Acc. } & \textbf{Parse} \\
        \midrule
        Gemma 3-27B             & 33.00 & 100.00    & 27.00 & 100.00    & 31.00 & 100.00     \\
        GLM-4.5V                & 31.00 & 98.00     & 21.00 & 98.00     & 28.00 & 96.00    \\
        Llama4-Scout 17B        & 18.00 & 100.00    & 16.00 & 100.00    & 25.00 & 100.00    \\
        Qwen2.5-VL-7B           & 19.00 & 100.00    & 18.00 & 100.00    & 26.00 & 100.00     \\
        Qwen2.5-VL-32B          & 28.00 & 100.00    & 27.00 & 100.00    & 26.00 & 100.00    \\
        Qwen3-Thinking 30B      & 23.00 & 86.00     & 19.00 & 98.00     & 32.00 & 77.00     \\
        Gemini 2.5 Pro          & 96.00 & 100.00    & 92.00 & 100.00    & 80.00 & 100.00    \\
        \bottomrule
    \end{tabular}}
    \caption{\textbf{Optimal Move Prediction (1 move from solved).} Top-1 accuracy and format compliance (Parse: \% outputs matching \texttt{ANSWER: [A-D]} or \texttt{<ANSWER>...</ANSWER>}) across prompt modalities.}
    \label{tab:solve-moves-main}
\end{table}
\subsection{Optimal Move Prediction}
\label{subsec:exp:move_prediction}

Table \ref{tab:solve-moves-main} shows Gemini 2.5 Pro excelling in all modalities, with perfect A-D parsing, indicating strong perception and Argmin selection (A in TEA). In contrast, most open-weights show 20-35\% with uneven modality behavior. Gemma-3-27B struggles in Image-only (27\%), reflecting poor face reconstruction and verification (Table~\ref{tab:combined-recon-verif}). Qwen-2.5-VL-7B performs better in Text-only (26\%) than Image-only (18\%) or Image+Text (19\%), suggesting noisy visual streams hinder fusion. Qwen-2.5-VL-32B maintains ~26-28\% with perfect parsing, indicating selection limits. Qwen3-VL-30B-A3B-Thinking has strong verification but weak format compliance (77\% parsing), affecting MCQ. GLM-4.5V shows good reconstruction and verification but low MCQ (31\% Image+Text; 21\% Image-only), highlighting selection/ranking issues over perception. All runs use the same number of items per modality for comparability.

\noindent\textbf{Takeaways:} (i) Static MCQ exposes a real selection gap: several models with decent reconstruction/verification still hover near chance because they mis-rank candidates or violate the A-D format; (ii) When vision is the bottleneck (e.g., Gemma-3, Qwen-7B), Image+Text performance can degrade compared to text-only (e.g., Llama 4); (iii) Strict parsing matters, non-A-D outputs directly convert to errors and can mask underlying competence; (iv) Only Gemini presents a consistent performance across tasks, high perceptual fidelity (reconstruction), strong cross-modal grounding (verification), and accurate selection (MCQ) while others are capped either by perception (i.e., in Gemma-3) or by the Argmin step (i.e., in GLM and Qwen2.5-32B).
\begin{table}[t]
    \centering
    \resizebox{0.8\columnwidth}{!}{%
    \begin{tabular}{@{}l c c c@{}}
        \toprule
        \textbf{Model} &  \textbf{$\Delta \uparrow$} & \textbf{EFR (\%) $\uparrow$} & \textbf{OTR (\%) $\downarrow$} \\
        \midrule
        Gemma 3-27B         & 0.00 & 22.00 & 78.00 \\
        GLM-4.5v            & -15.00 & 6.20 & 66.70 \\
        Llama4 Scout 17B    & +6.00 & 20.00 & 40.00 \\
        Qwen2.5-VL 7B       & +14.00 & 36.00 & 65.00\\
        Qwen2.5-VL 32B      & +11.00 & 34.71 & 64.00 \\
        Qwen3-Thinking 30B  & +11.00 & 27.81 & 50.00 \\
        \bottomrule
    \end{tabular}}
    \caption{Reflection-guided re-answering on Optimal Move Prediction (“Guided (Redacted)”). $\Delta$ is the absolute change in accuracy by applying reflection; EFR is the corrected share of initially wrong items; OTR is the share of correct items flipped to wrong. All runs use Image+Text modality, zero temperature, deterministic option shuffling, and strict A-D parsing.}
	\label{tab:reflection-guided}
\end{table}
\begin{table*}[t]
\footnotesize
    \centering
    \resizebox{0.75\textwidth}{!}{
    \begin{tabular}{l *{9}{c}}
        \toprule
        & \multicolumn{3}{c}{$\boldsymbol{d{=}1}$} & \multicolumn{3}{c}{$\boldsymbol{d{=}2}$} & \multicolumn{3}{c}{$\boldsymbol{d{=}3}$} \\
        \cmidrule(lr){2-4}\cmidrule(lr){5-7}\cmidrule(lr){8-10}
        \textbf{Model} & \textbf{Acc.} & \textbf{F1} & \boldmath$\kappa$
                       & \textbf{Acc.} & \textbf{F1} & \boldmath$\kappa$
                       & \textbf{Acc.} & \textbf{F1} & \boldmath$\kappa$ \\
        \midrule
        Gemma3                  & 32.00 & 0.30 & -0.025       & 31.5  & 0.309 & -0.027     & 27.00 & 0.262 & -0.099   \\
        GLM-4.5V                & 36.00 & 0.31 & 0.0014       & 30.00 & 0.25  & -0.0856    & 29.00 & 0.237 & -0.1011  \\
        Llama4 Scout 17B        & 28.50 & 0.27 & -0.084       & 27.75 & 0.27  & -0.079     & 27.25 & 0.27  & -0.084   \\
        Qwen2.5 VL 7B           & 37.00 & 0.21 & 0.0024       & 35.75 & 0.20  & 0.00059    & 36.5  & 0.20  & 0.00902  \\
        Qwen2.5 VL 32B          & 32.20 & 0.25 & -0.074       & 39.3 & 0.316 & 0.0403      & 41.25 & 0.335 & 0.0859  \\
        Qwen3-Thinking 30B      & 33.0  & 0.29 & -0.029       & 32.0 & 0.269 & -0.034      & 35.72 & 0.306 & 0.022       \\
        Gemini 2.5 Pro          & 79.0  & 0.78 & 0.680        & 65.0 & 0.632 & 0.475       & 62.0  & 0.606 & 0.439    \\
        \bottomrule
    \end{tabular}}
    \caption{Move-Effect results by depth ($d{=}1,2,3$). Higher is better unless noted.
    Macro-F1 over \textsc{decrease}, \textsc{no\_change}, \textsc{increase}.}
    \label{tab:move_effect_by_depth}
\end{table*}
\subsection{Reflection Guided Reasoning}
\label{subsec:exp:reflection}
As shown in Table~\ref{tab:reflection-guided}, the Qwen family achieves consistent, meaningful gains of +11-14 points (EFR $\approx$ 28-36\%), albeit with sizable OTR (50-65\%). Llama 4-Scout-17B:16E improves modestly (+6; EFR 20\%, OTR 40\%). Gemma-3-27B shows no net benefit (0), and GLM degrades ($-15$) due to high overthinking (OTR 66.7\% vs.\ EFR 6.2\%). These outcomes emphasize that label-free reflection can help, but is not uniformly positive; the EFR/OTR balance is decisive.

\noindent\textbf{Takeaways:}
Reflection under no-leak prompts can help, but gains are model-dependent: Qwen models consistently improve (+11, +14), Llama 4-Scout-17B:16E improves slightly (+6), Gemma-3-27B shows none, and some models overthink and regress. Net improvement hinges on the EFR-OTR trade-off i.e., raising EFR without inflating OTR, with GLM’s $-15$ explained by OTR 66.7\% overwhelming EFR 6.2\%. Finally, reasoning $\neq$ perception: these interventions strengthen selection on static items; they do not repair weak visual grounding or format compliance.

\subsection{Causal Move-Effect}
\label{subsec:exp:move_effect}
From Table \ref{tab:move_effect_by_depth}, the frontier model (Gemini 2.5 Pro) leads with consistently positive $\kappa$ and the strongest Macro-F1 across depths, indicating genuine causal forecasting beyond priors (though performance still drops as scramble depth increases). Open-source baselines cluster near chance. For Qwen-2.5-32B, micro-accuracy edges up with depth, but $\kappa$ shows what’s really happening: below chance at $d=1$ ($\approx -0.074$), barely above chance at $d=2$ ($\approx 0.040$), and only a small positive at $d=3$ ($\approx 0.086$). The confusion and prediction mix explain this: Qwen mostly predicts INCREASE and rarely predicts DECREASE. In practice, it treats most moves as “making things worse,” so it misses many true DECREASEs (low recall), keeping Macro-F1 modest ($\approx 0.25$ at $d=1$, $\approx 0.33$ at $d=3$) even when micro-accuracy improves. When the depth-wise class prior places more mass on INCREASE, this bias can inflate micro-accuracy, but $\kappa$ stays small/negative because it adjusts for the model’s skewed prediction mix, signaling little real evaluative skill. GLM shows a similar trend: chance-level $\kappa$ at $d=1$ declines  as depth increases, consistent with prior-chasing rather than a calibrated, causal assessment.

\noindent\textbf{Takeaways:} Chance-corrected $\kappa$ is the headline metric: it discounts depth-dependent priors and duplicated-option heuristics, so $\kappa > 0$ signals real pre-action understanding, while $\kappa \le 0$ indicates guessing or bias even when Micro looks acceptable. Only the frontier model demonstrates robust causal move-effect forecasting across depths (though still weaker at higher $d$); open-weights show shallow heuristics (INCREASE bias, DECREASE blind spot). We therefore emphasize $\kappa$ (with Micro, $q \cdot \pi$, and Macro-F1 for transparency) when comparing depth, and interpret Qwen-2.5-32B’s small $\kappa$ gains as prior-alignment effects more than substantive improvements in cube reasoning.

\subsection{Closed-Loop Control}
\label{subsec:exp:closed_loop}
\begin{figure*}[t]
    \centering
    \includegraphics[width=0.7\linewidth]{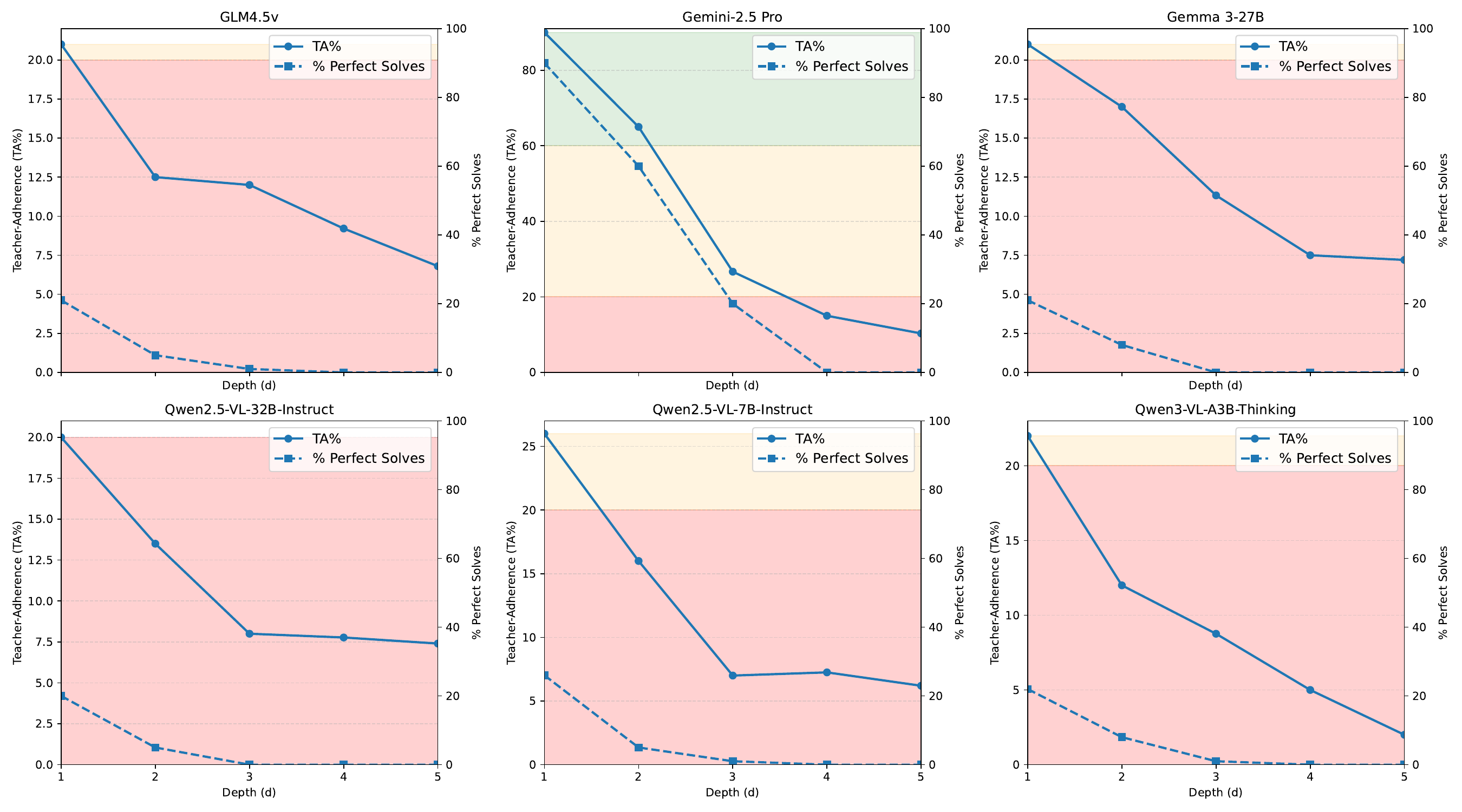}
    \caption{Step-by-step closed-loop results. \emph{TA\%} is Teacher-Adherence, i.e., $\frac{\text{avg correct steps}}{d}\times 100$. \emph{Perfect\%} is the share of episodes with perfect teacher adherence. Accuracies use \emph{unconditional} denominators (episodes that ended early count as incorrect at later steps).
    Bands (w.r.t TA\%): \colorbox{red!15}{0--20\%} \colorbox{orange!15}{20--60\%} \colorbox{green!12}{60--100\%}.}
    \label{fig:step-by-step-results}
\end{figure*}

Models degrade with depth at varying rates. From Fig. \ref{fig:step-by-step-results}, Gemini 2.5 Pro starts strong in the \colorbox{green!12}{green} band at \(d{=}1\), then declines with errors (from \(90\) to \(10\) TA\%). Open-weights cluster in the \colorbox{red!15}{red} band by \(d{\ge}3\): Qwen-2.5-VL-7B (from \(26\) to \(6\)), Qwen-2.5-VL-32B (from \(20\) to \(7.5\)), Qwen3-VL-30B (from \(22\) to \(9\)), GLM-4.5V (from \(20\) to \(9\)), and Gemma-3-27B (from \(20\) to \(7\)).

\noindent\textbf{Takeaways:} The results highlight several critical weaknesses in current models: (1) \textbf{Sequential Reasoning Collapses with Depth:} All models, regardless of size, show a severe collapse in performance as the number of sequential steps (depth $d$) increases. This demonstrates a fundamental fragility in tasks requiring long-horizon, closed-loop reasoning, (2) \textbf{Even Frontier Models are Brittle:} While Gemini 2.5 Pro is the clear leader, starting at 90\% Teacher-Adherence (TA) at $d=1$, its performance is not robust. It suffers a massive drop to 10\% TA, indicating that even the most advanced models are highly sensitive to sequential task length, (3) \textbf{Significant Gap Between Frontier and Open Models:} A stark performance divide exists. Gemini 2.5 Pro (the \colorbox{green!12}{green} band) starts with 3--4$\times$ higher adherence than the cluster of open-weight models (the \colorbox{red!15}{red} band), which only achieve 20--26\% TA at $d=1$ and fall to near-failure levels (6--9\% TA) by $d \ge 3$, and (4) \textbf{Planning, Not Just Perception, is the Bottleneck:} The analysis diagnoses \emph{why} models fail. Models with decent perception (like GLM-4.5V) still decline sharply, indicating the failure is not just in \emph{perception} but in the higher-level cognitive task of \emph{planning} and making correct sequential \emph{selections}.

\subsection{Learning-Curve / Recovery}
\label{subsec:exp:recovery}
\begin{table}[t]
    \centering
    \resizebox{\columnwidth}{!}{%
    \begin{tabular}{@{}l c c c c c c@{}}
        \toprule
    \multirow{2}{*}{\textbf{Model}} &
    \multirow{2}{*}{\textbf{SR}} &
    \multirow{2}{*}{\textbf{95\% CI}} &
    $\textbf{P}$ &
    $\textbf{P}$ &
    \textbf{Med@} &
    \textbf{Avg@} \\
    &  &  & $(1)$ & $(\le3)$ & \textbf{Solved} & \textbf{All} \\
        \midrule
        Gemma 3-27B             & 4.00  & [0.7, 19.5]       & 0.00  & 0.04  & 3.00  & 5.88    \\
        GLM-4.5v                & 2.00  & [0.0035, 0.10]    & 0.00  & 0.02  & 2.00  & 5.92    \\
        Llama4 Scout 17B        & 6.00  & [0.021, 0.162]    & 0.00  & 0.00  & 6.00  & 5.98    \\
        Qwen2.5 VL 7B           & 8.00  & [2.20, 25.00]     & 0.00  & 0.04  & 4.50  & 5.88    \\
        Qwen2.5 VL 32B          & 0.00  & [0.00, 13.30]     & 0.00  & 0.00  & 0.00  & 6.00    \\
        Gemini 2.5 Pro          & 40.00 & [11.76, 76.93]    & 0.00  & 0.40  & 3.00  & 4.8    \\
        \bottomrule
    \end{tabular}}
    \caption{Learning-curve leaderboard. 
    SR = solve rate; CI = 95\% Wilson interval; $P(1)$ and $P(\le3)$ are fractions solved within 1 and 3 attempts; 
    Med@Solved = median attempts (solved only); Avg@All = average attempts counting failures as max\_attempts.}
    \label{tab:learning_curve}
\end{table}

All episodes start at $d{=}3$; after the first deviation we allow six attempts. As shown in Table~\ref{tab:learning_curve}, Gemini 2.5 Pro leads (SR 40\%, $P(\le3)=0.40$, Med@Solved 3, Avg@All 4.8). Qwen 2.5-VL-7B is modest (SR 8\%, Med@Solved 4.5, Avg@All 5.88). Gemma 3-27B (SR 4\%, Med@Solved 3, Avg@All 5.88) and Llama 4-Scout-17B:16E (SR 6\%, Med@Solved 6, Avg@All 5.98) recover rarely; the latter typically only at the budget limit. GLM-4.5v is the lowest (SR 2\%, Med@Solved 2, Avg@All 5.92). Qwen 2.5-VL-32B does not recover within budget (SR 0\%, Avg@All 6.00). No system achieves immediate bounce-back ($P(1)=0$ for all). CIs are reported in Table~\ref{tab:learning_curve}.

\noindent\textbf{Takeaways:} Recovery from $d{=}3$ is non-trivial: most error-bearing episodes drift to the attempt limit rather than re-enter a solving trajectory. \textbf{Gemini 2.5 Pro} is comparatively strong yet still leaves 60\% of episodes unsolved within six attempts. The rest exhibit fragile local repair, \textbf{GLM-4.5v} is fast when it finds a fix (low $\text{Med@Solved}$) but almost never does (2\% SR), while \textbf{Llama 4-Scout-17B:16E} only succeeds at the ceiling (6-attempt median) with no early recoveries. Combined with adherence results, this suggests: (i) one-step skill does not translate into post-error control, (ii) immediate local repair is weak (no $P(1)$ successes), and (iii) effective recovery likely requires explicit mechanisms, short rollouts or self-checks before acting (TEA “Evaluation”), selective abstention/deferral, and probing deeper attempt budgets to locate saturation points.

\subsection{Diagnostics and Ablations}
\label{subsec:experiments:diagnostics}

We briefly probe two additional axes to check that our conclusions are not artefacts of a particular protocol choice: abstention-aware control in the step-by-step test, and reflection regimes with vs.\ without label leakage. We also summarize fairness diagnostics for the Move-Effect sampler.

\noindent\textbf{Abstention (Selective Control).}
In the closed-loop Step-by-Step test, we optionally allow an explicit  ``I don't know (IDK)'' action, mapped to a no-operation policy in the environment \cite{kalai2025languagemodelshallucinate} (teacher\_on\_abstain: apply the teacher move so the episode length is preserved). Primary metrics (TA\% and Perfect-Solves\%) remain unchanged and keep unconditional denominators: IDK receives no credit and counts as incorrect when computing adherence. Abstention can therefore prevent a bad move but cannot inflate TA\%.

To understand how models use abstention, we additionally report coverage (answered / total), selective accuracy (correct / answered), and APA, an abstention-penalized accuracy with weight $\lambda = 0.25$ on IDK. Table~\ref{tab:abstention} shows a small slice of this analysis at $d=5$.

\begin{table}[t]
    \centering
    \footnotesize
    \resizebox{0.8\columnwidth}{!}{%
    \begin{tabular}{lcccc}
    \toprule
    \multirow{2}{*}{\textbf{Model}} &
    \textbf{Coverage} & 
    \textbf{Sel.} &
    \textbf{IDK} &
    \textbf{APA} \\
    & (\%) &  Acc. & (\%) & ($\lambda=0.25$) \\
    \midrule
    Qwen2.5-VL-32B   & 99.3  & 23.7 & 0.0  & 0.24 \\
    Gemma 3-27B      & 100.0 & 17.3 & 0.0  & 0.17 \\
    Gemini 2.5 Pro   & 87.5  & 28.6 & 12.5 & 0.28 \\
    \bottomrule
\end{tabular}}
    \caption{Abstention-aware selective metrics on the Step-by-Step test with IDK enabled and teacher\_on\_abstain at $d=5$. Coverage is answered / total; Sel.\ Acc is correct / answered; APA is an abstention-penalised accuracy.}
    \label{tab:abstention}
\end{table}

Gemini shows the highest precision when it chooses to act (Sel.\ Acc $28.6\%$) but at the cost of lower coverage ($87.5\%$) driven by explicit abstention (IDK $12.5\%$); with $\lambda = 0.25$ this yields the best penalty-aware score (APA $0.28$). Qwen2.5-VL-32B and Gemma-3 almost never abstain (IDK $0\%$), maintain near-total coverage ($99.3$-$100\%$), but achieve lower selective accuracy ($23.7\%$ and $17.3\%$) and lower APA ($0.24$ and $0.17$). Overall, calibrated abstention can modestly improve penalty-aware accuracy by avoiding low-confidence moves, but it does not change the depth-collapse pattern in TA\% or Perfect-Solves\%.

\noindent\textbf{Reflection regimes: Unredacted vs.\ Redacted.}
Table~\ref{tab:reflection-guided} reports the label-safe ``Guided (Redacted)'' regime, where reflection operates without access to the gold answer. For completeness, we also evaluated a leaky ``Guided (Unredacted)'' regime in which the reflection prompt includes the correct option as an oracle hint (upper-bound oracle guidance). Under Unredacted, every model improves sharply: for example, GLM-4.5v and Llama 4-Scout-17B:16E climb from $31\%$ and $18\%$ initial accuracy to $100\%$ (gains of $+69$ and $+82$ points), and Qwen models jump to $59$-$72\%$ final accuracy with gains of $+42$-$+50$ points. Gemma 3-27B rises from $22\%$ to $81\%$.

These results frame the Redacted curves as realistic, label-free behavior and the Unredacted curves as upper bounds. In the leaky regime, reflection can make almost any model look strong on static MCQ; in the label-safe regime, gains are smaller and strongly model-dependent (Qwen models benefit, Gemma does not, GLM degrades), consistent with the EFR/OTR trade-off in~\S\ref{subsec:exp:reflection}. This supports our decision to headline Redacted results in the main experiments and treat Unredacted as an oracle ablation.

\subsection{Headline Summary}
\label{subsec:exp:headline}
\begin{figure}[t]
    \centering
    \includegraphics[width=0.8\columnwidth]{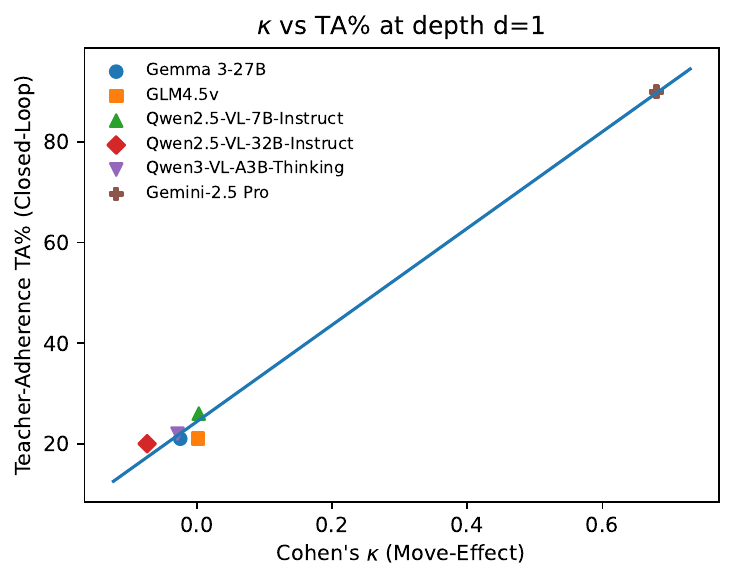}
    \caption{Move-Effect $\kappa$ vs. Closed-Loop TA\% at depth $d{=}1$.
    Pearson $r=0.997$ (95\% CI [0.972, 1.000]); least-squares fit shown.
    The relation weakens among models with $\kappa\!\approx\!0$.}
    \label{fig:kappa-ta-d1}
\end{figure}

At fixed difficulty, \emph{pre-action causal evaluation} (Move-Effect $\kappa$) tracks \emph{closed-loop control} (TA\%, Perfect-Solve\%) and \emph{post-error recoverability} (SR).
At $d=1$ the association is very strong (Fig.~\ref{fig:kappa-ta-d1}: $r=0.997$). The only model with consistently positive $\kappa$ across depths, e.g., Gemini 2.5 Pro (Table~\ref{tab:move_effect_by_depth}), also leads \emph{in Step-by-Step} closed-loop metrics (TA\%, Perfect-Solve\%) at shallow scrambles (Fig.~\ref{fig:step-by-step-results}) and is the only system to show a meaningful \emph{Learning Curve} (Table~\ref{tab:learning_curve}), where we grant a retry budget and continue the episode even when a non-optimal move is chosen to test recovery.
However, once outputs feed back into inputs, performance \emph{degrades sharply with depth} (Fig.~\ref{fig:step-by-step-results}); even the best model falls steeply and early recovery (\mbox{$P(1)$}, $P(\le3)$) is rare.
Among open-source models, $\kappa$ clusters near $0$, so the $\kappa \rightarrow$TA relation weakens at $d{=}2,3$ (per-depth plots in \emph{Supplementary Material}.).

\section{Conclusion}
\label{section:conclusion}
We introduced Cube Bench, a compact, simulator-based benchmark that uses the Rubik’s Cube as a controlled testbed to probe the full \emph{see $\rightarrow$ evaluate $\rightarrow$ act $\rightarrow$ reflect $\rightarrow$ recover} steps in multimodal language models. Across seven tests, we showed that current systems can reconstruct and verify the state with high accuracy, yet struggle to turn local decisions into stable sequence of decisions: depth increases quickly lead to collapse in performance, post-error recovery is rare, and strong one-step skills often fail to translate into robust long-horizon behavior. Pre-action causal evaluation (Move-Effect $\kappa$) tracks both adherence and recovery, highlighting the value of decision-aware metrics beyond raw accuracy. Reflection can help, but only under label-safe regimes and with careful control of overthinking. Taken together, these results suggest that improving multimodal LLMs will require not only better perception, but also explicit mechanisms for pre-action evaluation, selective control, and recovery. Cube Bench offers a lightweight, reproducible platform for studying these issues in isolation and for comparing future models and intervention strategies under matched conditions.

\section{Limitations}
\label{section:limitations}
Our work on Cube Bench has some limitations. It uses Rubik's Cube to test spatial perception and short-horizon planning, but the results may not apply to broader tasks like robotics or web-agent tasks. Due to poor performance of existing MLLMs, our evaluations are limited to shallow horizons. Nonetheless, our framework could produce arbitrarily long-horizon tasks for future studies. Tasks use low scramble depths and multiple-choice formats, limiting insights into complex \emph{thinking and reasoning} errors. 

\section*{Acknowledgment}
\label{section:acknowledgment}
We are very grateful to Ivan Vulić and Chengzu Li for their thoughtful and constructive feedback on an early version of the Cube benchmark idea.

{
    \small
    \bibliographystyle{ieeenat_fullname}
    \bibliography{main}
}


\onecolumn
\appendix
\tcbset{
  graynote/.style={
    enhanced,
    colframe=black!100,
    colback=green!10,
    title filled,
    colbacktitle=green!25,
    coltitle=black,
    fonttitle=\bfseries\small,
    boxrule=0.6pt,
    arc=2pt,
    left=4pt, right=4pt, top=4pt, bottom=4pt,
    width=\columnwidth,
    fontupper=\small\ttfamily,
    breakable
  }
}

\newenvironment{systemprompt}
  {\begin{tcolorbox}[graynote,title=System Prompt]}
  {\end{tcolorbox}}

\newenvironment{userprompt}
  {\begin{tcolorbox}[graynote,title=User Prompt]}
  {\end{tcolorbox}}

\clearpage
\section{\texttt{VirtualCube} Implementation Details}
\label{app:virtualcube}

\subsection{Visual Rendering}
\label{app:virtualcube:visual_render}
\begin{figure}[t]
    \centering
    \begin{subfigure}[b]{0.48\linewidth}
        \centering
        \includegraphics[width=\linewidth]{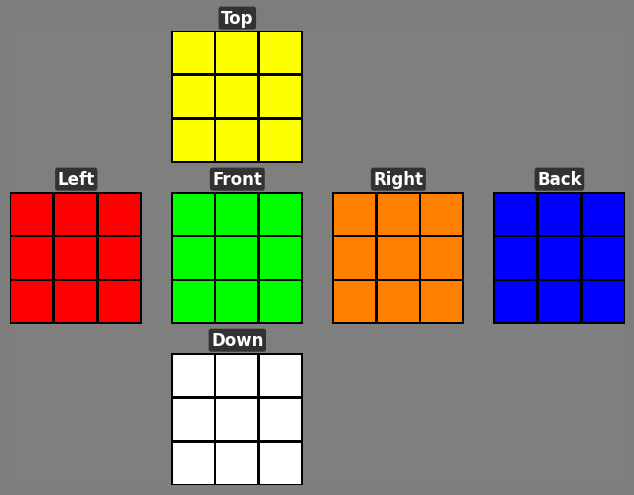}
        \caption{A Rubik's Cube in its solved state.}
        \label{fig:cube_solved}
    \end{subfigure}
    \hfill
    \begin{subfigure}[b]{0.48\linewidth}
        \centering
        \includegraphics[width=\linewidth]{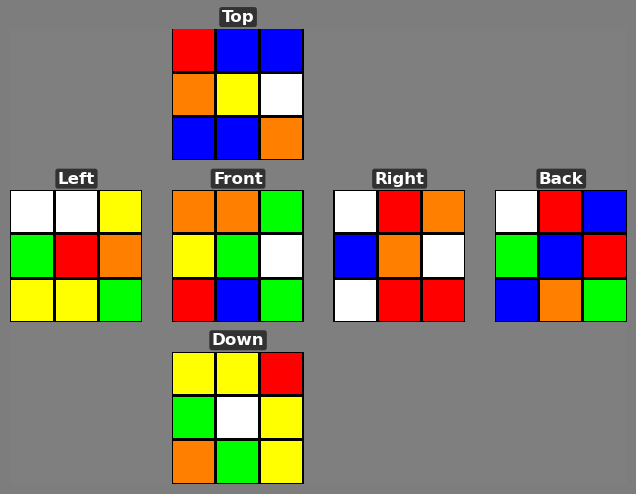}
        \caption{A Rubik's Cube in a scrambled state (e.g., at a given depth).}
        \label{fig:cube_scramble}
    \end{subfigure}
    \caption{\textbf{Example Rubik's Cube States.} The left image shows a cube in its fully solved configuration, where each face consists of a single color. The right image displays a cube in a scrambled state, typical of a starting position in a puzzle-solving task.}
    \label{fig:combined_cube_states}
\end{figure}

To ensure complete state observability and eliminate ambiguity caused by occlusion in 3D perspective views, we utilize a standardized \textbf{unfolded ``cube net'' visualization} for all image-based tasks. As illustrated in Figure~\ref{fig:combined_cube_states}, this rendering style flattens the cube's state into a 2D plane, exposing all six faces simultaneously: \textbf{Top}, \textbf{Left}, \textbf{Front}, \textbf{Right}, \textbf{Back}, and \textbf{Down}. Explicit text labels are rendered above each face to aid model grounding and ensure that the orientation is unambiguous. 

The background is rendered in a uniform, \textbf{neutral gray}. This design choice maximizes contrast against the vibrant sticker colors (W, Y, R, O, G, B) and eliminates potential visual confounders, ensuring that the model's attention is focused exclusively on the state configuration rather than background artifacts.

\subsection{Oracle Distance Implementation Details}
\label{app:virtualcube:oracle}
To ensure reproducible and consistent evaluation of model performance, we utilize a deterministic oracle to compute the exact distance-to-solved $d(s)$ for every state.

\subsubsection{Solver Algorithm and Optimality}
The oracle distance $d(s)$ is defined as the length of the strictly shortest path from the current state $s$ to the solved state in the Face Turn Metric (FTM).

To compute this, we utilize the \texttt{RubikOptimal} Python library~\cite{rubikoptimal}. Unlike heuristic solvers (e.g., standard Two-Phase algorithms) which may return suboptimal solution lengths, \texttt{RubikOptimal} implements Korf's IDA* (Iterative Deepening A*) algorithm~\cite{korf1997finding} with pruning tables. This guarantees that for any state $s$, the returned distance is the theoretical minimum (God's Number):

\[
    d(s) = \min \{ |P| : P \text{ transforms } s \to s_{\text{solved}} \}
\]
where $|P|$ denotes the path length in FTM.

\subsubsection{Metric Definition (Face Turn Metric)}
The benchmark strictly adheres to the Face Turn Metric (HTM/FTM). The solver is configured to treat the following 18 moves as atomic actions with a cost of 1:
\begin{itemize}
    \item \textbf{Clockwise:} $U, D, F, B, L, R$
    \item \textbf{Counter-Clockwise:} $U', D', F', B', L', R'$
    \item \textbf{Double Turns:} $U2, D2, F2, B2, L2, R2$
\end{itemize}
Slice moves ($M, E, S$) and wide moves ($u, d, f$, etc.) are not permitted as atomic actions and are decomposed into their constituent face turns for distance calculation.

\subsubsection{Progress Computation}
In the \textit{Closed-Loop Step-by-Step} task, the oracle evaluates the transition from state $s_t$ to $s_{t+1}$ caused by a model's chosen action $a_t$. We determine if a step constitutes progress by comparing the exact oracle distances:

\[
    \text{IsOptimal}(a_t) \iff d(s_{t+1}) < d(s_t)
\]

Since $d(s)$ is guaranteed optimal, any move satisfying this condition is mathematically proven to be on a shortest path to the solution.

\subsection{Reproducibility and Seed Lists}
To ensure exact reproducibility across all models and runs, we strictly enforce a deterministic generation protocol. For every episode $i$ at depth $d$, the simulator is initialized with \texttt{seed = i}. This determines the scramble state $s_{d,i}$, the teacher inverse plan, and the distractor options.

We provide the full metadata for these episodes in the supplementary code package (file: \texttt{configs/seed\_lists.json}). This file explicitly maps every episode index to its corresponding scramble string and ground-truth attributes. Future benchmarks can strictly replicate our test set by loading these pre-generated configurations or by re-running the generator with the same seed integers.

\clearpage
\section{Detailed Task Specifications and Fairness Control}
\label{app:task_specs}
\subsection{Cube Face Reconstruction}

\subsubsection{Definition}
\label{app:def:cfr}

Let $I$ be the input image and $M^\star \in \mathcal{V}^{3\times3}$ the ground-truth face matrix over vocabulary $\mathcal{V}$.
The model outputs $\widehat{M}=f(I)$ after normalization $N(\cdot)$.
Element-wise accuracy is
\[
\text{Acc}_{\text{elem}}=\frac{1}{9}\sum_{i=1}^{3}\sum_{j=1}^{3}\mathbf{1}\{N(\widehat{M}_{ij})=M^\star_{ij}\},
\]
and matrix accuracy is $\text{Acc}_{\text{mat}}=\mathbf{1}\{N(\widehat{M})=M^\star\}$.
Parse-violations are counted when $N(\widehat{M}) \notin \mathcal{V}^{3\times3}$ (wrong shape/tokens).

\subsubsection{Generation \& Fairness Controls}
\label{app:ctrl:cfr}

Images are generated with \texttt{VirtualCube} using a fixed camera pose, a standard color scheme, and uniform random cube states sampled by scramble depth $d$, ensuring that color-token priors do not deviate from a uniform distribution by more than 5\% (maximum deviation $<$ 0.05).. We disable motion blur and heavy lighting effects and clamp minor rotation jitter to reduce nuisance variance.

\subsection{Cross-Modal Verification}

\subsubsection{Definition}
\label{app:def:verification}

Let $I$ be the cube-net image, $h$ the front-face hypothesis, and $y\!\in\!\{\textsc{Yes},\textsc{No}\}$ the truth label computed from the rendering state. The model outputs $\hat{y}\!\in\!\{\textsc{Yes},\textsc{No}\}$ after normalization. Balanced accuracy:

\begin{align*}
    \text{BAcc}=\tfrac{1}{2}\big(\Pr[\hat{y}=\textsc{Yes}\mid y=\textsc{Yes}] + \\
    \Pr[\hat{y}=\textsc{No}\mid y=\textsc{No}]\big).
\end{align*}

Parse-violations occur when the response lacks a valid \texttt{Yes/No} tag.

\subsubsection{Generation \& Fairness Controls}
\label{app:ctrl:verification}

Hypotheses are auto-derived from ground-truth states produced by \texttt{VirtualCube}. We enforce a $50/50$ Yes/No prior per depth (and per template, when applicable), include polarity reversals (e.g., “is green”/“is not green”), and randomize surface forms to avoid lexical shortcuts. Ambiguous references are excluded.

\subsection{Optimal Move Prediction (Static MCQ)}

\subsubsection{Definition}
\label{app:def:prediction}

Let $T(s,a)$ be the transition and $d(\cdot)$ the fixed distance (FTM/HTM). For a one-move scramble $s=T(s^\star,a_s)$,
\[
y=\arg\min_{a\in\{a_A,a_B,a_C,a_D\}} d\!\left(T(s,a)\right)
\]
\[
\text{and at }d{=}1\text{, }y=a_s^{-1}.
\]
A prediction $\hat{y}\in\{A,B,C,D\}$ is correct iff $\hat{y}=y$.

\subsubsection{Generation \& Fairness Controls}
\label{app:ctrl:prediction}

We fix \verb|n_moves=1| so the gold action is the inverse of the scramble (\emph{unique-optimal} under the chosen metric). Camera pose/colors match Recognition/Verification; the text serialization reuses the same normalization. Options are full-entropy shuffled; over many items this approximates uniform A-D placement.

\subsection{Reflection-Guided Reasoning}

\subsubsection{Definition}
\label{app:def:refl}

Let $I = \{1, 2, \dots, N\}$ be the set of $N$ test items.
For each item $i \in I$:
\begin{itemize}
    \item $s_i$ is the sample data (cube state, options A, B, C, D).
    \item $g_i \in \{A, B, C, D\}$ is the ground-truth correct (gold) answer.
    \item $\hat{y}_{i,1}$ is the initial "Draft" choice from the model.
    \item $\hat{y}_{i,2}$ is the final "Re-answer" choice from the model, after processing reflection $r_i$.
    \item $1\{\cdot\}$ is the indicator function.
\end{itemize}

\noindent We define the initial accuracy (score) $a_{i,1}$ and final accuracy $a_{i,2}$ for item $i$:
\[
    a_{i,1} = 1\{\hat{y}_{i,1} = g_i\} \\
    a_{i,2} = 1\{\hat{y}_{i,2} = g_i\}
\]

\noindent To define EFR and OTR, we partition the total set $I$ based on initial performance:
\begin{itemize}
    \item \textbf{Initially Wrong Set ($I_W$):} $I_W = \{i \in I \mid a_{i,1} = 0\}$. Let $N_W = |I_W|$.
    \item \textbf{Initially Correct Set ($I_C$):} $I_C = \{i \in I \mid a_{i,1} = 1\}$. Let $N_C = |I_C|$.
\end{itemize}
(Note: $N_W + N_C = N$).

\paragraph{Initial Acc.}
The accuracy before reflection (from the "Draft" phase)[cite: 196].
\[
    \text{Initial Acc} = \frac{1}{N} \sum_{i=1}^{N} a_{i,1}
\]

\paragraph{Final Acc (All).}
The accuracy after reflection and re-answering, measured over all $N$ items[cite: 196].
\[
    \text{Final Acc (All)} = \frac{1}{N} \sum_{i=1}^{N} a_{i,2}
\]

\paragraph{Gain ($\Delta$).}
The net gain in accuracy after reflection[cite: 197].
\[
    \Delta = \text{Final Acc (All)} - \text{Initial Acc}
\]

\paragraph{Error-Fix Rate (EFR).}
The fraction of previously incorrect items ($I_W$) that were corrected in the "Re-answer" phase[cite: 197].
\[
    \text{EFR} = 
    \begin{cases} 
        \frac{1}{N_W} \sum_{i \in I_W} a_{i,2} & \text{if } N_W > 0 \\ 
        \text{NaN} & \text{if } N_W = 0 
    \end{cases}
\]

\paragraph{Overthink Rate (OTR).}
The fraction of previously correct items ($I_C$) that were flipped to wrong in the "Re-answer" phase[cite: 198].
\[
    \text{OTR} = 
    \begin{cases} 
        \frac{1}{N_C} \sum_{i \in I_C} (1 - a_{i,2}) & \text{if } N_C > 0 \\ 
        \text{NaN} & \text{if } N_C = 0 
    \end{cases}
\]

\subsubsection{Generation \& Fairness Controls}
\label{app:ctrl:refl}

Items are rebuilt deterministically via per-index seeds; A-D options are fully shuffled; the FTM/HTM
distance metric is fixed per run.
The re-answer prompt defines the deterministic tie-break (A $>$ B $>$ C $>$ D) and the strict output format.
We log tokens and latency for both reflection and re-answer.

\subsection{Closed-Loop Step-by-Step}

\subsubsection{Definition}
\label{app:def:stepbystep}
Let $s_0$ be the depth-$d$ start state, $T$ the transition, and $d(\cdot)$ the fixed metric. The teacher sequence $(a^{\text{teach}}_1,\dots,a^{\text{teach}}_d)$ is the inverse of the scramble. At step $t$ the model outputs $a_t\!\in\!\{a_A,a_B,a_C,a_D\}$ (or \texttt{IDK}), the environment updates $s_t=T(s_{t-1}, a_t^\star)$ per policy, and $\Delta_t=d(s_t)-d(s_{t-1})$. The episode halts at the first $t$ with $\Delta_t\!\ge\!0$ or a parse violation; otherwise it runs up to $d$ steps.

\subsubsection{Generation \& Fairness Controls}
\label{app:ctrl:stepbystep}

States come from \texttt{VirtualCube} with depth $d$. At each step we (i) compute the teacher move from the reversed scramble, (ii) draw 3 unique distractors from the legal move set, and (iii) \emph{full-entropy} shuffle options A-D to dilute position bias. The distance metric (HTM/FTM) is fixed per run. If the parsed option is teacher or oracle-optimal, we apply it; if parsing fails, we apply a deterministic fallback (A) for progress checking and halt on non-progress.

\subsection{Abstention (Selective Control)}
\label{app:abstention}
\subsubsection{Definition}
\label{app:def:abstention}

We extend the closed-loop next-move task to allow an explicit abstention token \texttt{IDK}. Let $s_0$ be the depth-$d$ start state, $d(\cdot)$ a fixed distance metric (FTM/HTM), and $(a_1^{\text{teach}},\dots,a_d^{\text{teach}})$ the teacher sequence (inverse scramble). At step $t$, the model outputs $y_t\!\in\!\{\text{A},\text{B},\text{C},\text{D},\texttt{IDK}\}$; the option letter maps to a concrete move $a_t\in\mathcal{M}$. The environment updates per the abstention \emph{policy} (below), yielding $s_t$, and we define the per-step progress
\[
\Delta_t \;=\; d(s_t) - d(s_{t-1}).
\]
The episode halts at the first $t$ with (i) non-progress, $\Delta_t\!\ge\!0$, or (ii) a parse violation (no valid token). Otherwise it runs to $d$.

\paragraph{Primary scoring.}
\emph{Teacher Adherence (TA\%)} and \emph{Perfect\%} ignore abstention for credit: \texttt{IDK} receives no credit (counts as incorrect) and cannot contribute to a Perfect episode. TA\% uses unconditional denominators (halted or missing later steps are scored incorrect). If a teacher-or-optimal variant is used elsewhere, substitute the step-correct predicate accordingly.

\paragraph{APA (Abstention-Penalized Accuracy).}
APA assigns partial credit $\lambda\!\in\![0,1]$ to abstentions (\texttt{IDK}) and full credit $1$ to correct answers, with wrong answers receiving $0$:
\[
\mathrm{APA}_\lambda=\frac{n_{\text{correct}}+\lambda\,n_{\text{IDK}}}{n_{\text{total}}}.
\]
Here $n_{\text{total}}$ counts all decisions (including parse failures), $n_{\text{IDK}}$ counts abstentions, and $n_{\text{correct}}$ counts correct answered moves. Properties: (i) $\mathrm{APA}_0$ reduces to standard micro-accuracy that treats \texttt{IDK} as incorrect; (ii) $\mathrm{APA}_1 = 1 - \frac{n_{\text{wrong}}}{n_{\text{total}}}$ (abstentions incur no penalty); (iii) it decomposes via coverage ($c=\tfrac{n_{\text{answered}}}{n_{\text{total}}}$) and selective accuracy ($\text{sel-acc}=\tfrac{n_{\text{correct}}}{n_{\text{answered}}}$) as
\[
\mathrm{APA}_\lambda \;=\; c\cdot \text{sel-acc} \;+\; (1-c)\cdot \lambda.
\]
Unless stated otherwise, we set $\lambda{=}0.25$.

\paragraph{Selective metrics.}
Abstention is evaluated separately via coverage and abstention-aware metrics:
\[
\text{coverage} = \frac{n_{\text{answered}}}{n_{\text{total}}},\qquad
\text{sel-acc} = \frac{n_{\text{correct}}}{n_{\text{answered}}},\qquad
\text{APA}_\lambda = \frac{n_{\text{correct}} + \lambda\, n_{\text{IDK}}}{n_{\text{total}}}.
\]
Here $n_{\text{answered}}$ counts \{A,B,C,D\} responses; $n_{\text{IDK}}$ counts abstentions; $n_{\text{total}}$ counts all decisions (including parse failures). Unless stated, we use $\lambda{=}0.25$.

\paragraph{Parsing \& error model.}
We accept \verb|<ANSWER> IDK </ANSWER>|, \verb|ANSWER: IDK|/\verb|E|, and the phrase “I don’t know” (case-insensitive). Any other malformed output is a parse failure (\texttt{MISSING}) and is treated as an incorrect answer for TA\%, contributes to $n_{\text{total}}$, and halts the episode.

\paragraph{Policies on abstain.}
We support two deterministic policies:
\begin{itemize}
  \item \texttt{teacher\_on\_abstain} (default): apply $a_t^{\text{teach}}$ when $y_t{=}\texttt{IDK}$ to preserve episode length; no credit is awarded.
  \item \texttt{skip\_item}: halt the episode immediately when $y_t{=}\texttt{IDK}$.
\end{itemize}
Both policies preserve the unconditional-denominator convention for TA\%/Perfect\%.

\subsubsection{Generation \& Fairness Controls}
\label{app:ctrl:abstention}

\paragraph{Prompting.} When abstention is enabled, the system prompt adds \texttt{IDK} as a fifth option and enforces a strict output tag (\verb|<ANSWER> X </ANSWER>|). Tie-breaking and decision rules are otherwise unchanged.

\paragraph{Decoding.} Greedy decoding (\texttt{temperature=0.0}) with a hard cap on new tokens (sufficiently large to avoid truncation $\approx 2^{16}$). Outputs are case-folded and whitespace-stripped before parsing.

\paragraph{Option set construction.} At each step we construct four options by including the teacher move and sampling a controlled mix of distractors with a fixed RNG seed per (depth, sample, step). Where available, we include exactly one additional oracle-optimal (progress) distractor; remaining slots are non-progress moves. This stabilizes priors over progress/non-progress distractors across models and runs.

\paragraph{Randomness control.} We use a deterministic seed derived from $(d,\text{sample\_id},t)$ for option shuffling and distractor selection, ensuring identical option sets for all models.

\paragraph{Confidence gates (optional).} A global confidence threshold can be specified; if a model’s self-estimated confidence drops below the threshold, \texttt{IDK} is permitted. Thresholds are fixed per run and reported.

\paragraph{Logged configuration.} We log (\textit{i}) policy (\texttt{teacher\_on\_abstain} or \texttt{skip\_item}), (\textit{ii}) $\lambda$ for APA, (\textit{iii}) any confidence threshold, (\textit{iv}) coverage by step, and (\textit{v}) selective counts $(n_{\text{correct}},n_{\text{wrong}},n_{\text{IDK}},n_{\text{total}})$.

\subsubsection{Results}
\label{app:results:abstention}

\begin{table}[t]
    \centering
    \begin{tabular}{@{}l c c c c@{}}
        \toprule
        \textbf{Model} & \textbf{Coverage \%} & \textbf{Sel. Acc \%} & \textbf{IDK \%} & \textbf{APA (s{=}0.25)} \\
        \midrule
        Qwen2.5\,-VL\,32B      & 99.29 & 23.74 & 0.00 & 0.24 \\
        Gemma 3\,-27B          & 100.00 & 17.32 & 0.00 & 0.17 \\
        Gemini 2.5\,Pro        & 87.50 & 28.57 & 12.50 & 0.28 \\
        \bottomrule
    \end{tabular}
    \caption{Abstention-aware selective metrics for \textsc{StepByStepTest} with IDK enabled and teacher on abstain at $d=5$.}
    \label{tab:sbs_selective_macro}
\end{table}

\paragraph{Brief findings.}
Gemini 2.5-Pro shows the highest precision when it chooses to act (\emph{Sel.~Acc} 28.6\%) but at the cost of lower coverage (87.5\%) driven by explicit abstention (\texttt{IDK} 12.5\%); with $\lambda{=}0.25$ this yields the best penalty-aware score (APA 0.28). Qwen2.5-VL-32B and Gemma 3-27B almost never abstain (IDK 0\%), maintain near-total coverage (99.3-100\%), but achieve lower selective accuracy (23.7\% and 17.3\%), resulting in lower APA (0.24 and 0.17). Overall, this suggests a trade-off: calibrated abstention can modestly improve penalty-aware accuracy by avoiding low-confidence moves, whereas always answering preserves coverage but depresses precision. \emph{Note:} Gemini’s row is from a smaller, deeper run (5-move scramble, few episodes), so variance is higher and comparability is limited.

\subsection{Causal Move-Effect Probe}

\subsubsection{Definition}
\label{app:def:moveeffect}

Let $d(\cdot)$ be fixed (FTM via Optimal Oracle i.e., IDA* + pattern match heuristics), $T$ the transition. For any move $a$:
\[
\Delta(a)=d(T(s,a))-d(s),\quad
\]
\[
y(a)=
\begin{cases}
    \textsc{DEC} & \Delta(a)<0 \text{ or } T(s,a)\text{ solves},\\
    \textsc{NC}  & \Delta(a)=0,\\
    \textsc{INC} & \Delta(a)>0.
\end{cases}
\]
Given options $\{a_A,a_B,a_C,a_D\}$, the model outputs $\hat y_K\in\{\textsc{DEC},\textsc{NC},\textsc{INC}\}$ for each $K\in\{A,B,C,D\}$. Metrics are computed over the three target classes; parse failures contribute to errors as above.

\subsubsection{Generation \& Fairness Controls}
\label{app:ctrl:moveeffect}

\paragraph{Option construction (A--D) with fairness.}
To avoid exploitable priors and slot biases, each item’s four options are built as:
\begin{enumerate}
    \item \textbf{One-each + feasible double.} Prefer one move from each class present (\textsc{DEC}/\textsc{NC}/\textsc{INC}) and add a \emph{double} from a class that has at least two distinct neighbors.
    \item \textbf{Depth-aware target mix.} For depth $d$, estimate the feasibility of doubling class $c$ as
    
    $$
    \hat{p}_2(c\mid d)
    = \frac{\displaystyle \sum_{i:\, d_i = d} \mathbf{1}\!\left\{\,\lvert \mathrm{bucket}^{(i)}_{c}\rvert \ge 2 \,\right\} + \alpha}
             {\displaystyle \sum_{i:\, d_i = d} 1 \;+\; 2\alpha}
    \quad\text{with }\alpha=1.
    $$

    The per-depth \emph{target mix} is $0.25$ for each class (one slot) plus the extra $0.25$ allocated proportionally to $\hat{p}_2(\cdot\mid d)$; we pick the doubled class that is most underrepresented vs.\ this target.
    \item \textbf{Slot balancing.} The slot holding the doubled class rotates (A$\!\rightarrow$B$\!\rightarrow$C$\!\rightarrow$D), and the remaining moves are greedily assigned to flatten historical slot$\times$class frequencies.
\end{enumerate}

\subsection{Learning Curve (Closed-Loop MCQ)}
\subsubsection{Definition}\label{app:def:lcurve}

Let \(\mathcal{S}\) be cube states, \(\mathcal{M}\) the standard move set (quarter/half turns in Singmaster), and
\(m(s)\) the deterministic transition obtained by applying \(m\in\mathcal{M}\) to \(s\in\mathcal{S}\).
A fixed oracle distance \(d:\mathcal{S}\to\mathbb{N}_0\) (FTM/HTM; chosen once per run) gives the
\emph{distance-to-solved}. The start state \(s_0\) is produced by a depth-\(d\) scramble with
per-episode seed; the \emph{teacher plan} \(\pi_0\) is the inverse scramble
(shortest plan from \(s_0\) to solved), with head \(m^\star\).

\paragraph{Progress-making moves.}
For state \(s\), the progress set is
\[
G(s)\;=\;\{\,m\in\mathcal{M}\;:\; d(m(s))<d(s)\ \text{ or }\ m(s)\ \text{is solved} \,\}.
\]
An option is \emph{progress-making} iff it is in \(G(s)\).

\paragraph{MCQ construction.}
At attempt \(t\), from state \(s_t\) we construct an MCQ with four distinct options:
one option is sampled from \(G(s_t)\) (when \(G(s_t) \neq \varnothing\)), and the remaining
three are sampled from \(\mathcal{M}\setminus G(s_t)\) when feasible; options are then
uniformly shuffled into A-D using an RNG independent of the scramble seed.
(If \(|\mathcal{M}\setminus G(s_t)|<3\), a rare fallback samples without the
progress/non-progress restriction; evaluation remains defined.)

\paragraph{Parsing and attempts.}
The model outputs a single letter \(X_t\in\{\text{A},\text{B},\text{C},\text{D}\}\) using either
\verb|<ANSWER>X</ANSWER>| or \verb|ANSWER: X| (case-insensitive).
A \emph{parse-fail} is recorded if neither pattern is found.
An \emph{attempt} is counted for every query (including parse-fails).

\paragraph{Accept-progress control policy.}
Let \(m^\star\) be the head of \(\pi_t\) (teacher next move), and let \(a_t\) be the concrete move
mapped from the chosen option \(X_t\) (undefined on parse-fail).
The environment update is:
\[
(s_{t+1},\pi_{t+1}) \;=\;
\begin{cases}
\bigl(m^\star(s_t),\ \mathrm{tail}(\pi_t)\bigr), & \text{if } a_t=m^\star \quad\text{(apply teacher head)},\\[4pt]
\bigl(a_t(s_t),\ \mathrm{Solve}(a_t(s_t))\bigr), & \text{if } a_t\in G(s_t) \quad\text{(apply \& replan)},\\[4pt]
\bigl(a_t(s_t),\ \langle a_t^{-1}\rangle \,\Vert\, \pi_t\bigr), & \text{if } a_t\notin G(s_t) \quad\text{(apply \& push inverse)},\\[4pt]
\bigl(s_t,\ \pi_t\bigr), & \text{if parse-fail (state unchanged)}.
\end{cases}
\]
Here, \(\mathrm{Solve}(\cdot)\) is the oracle shortest-plan solver (returns an empty plan at solved),
\(\langle a_t^{-1}\rangle\) is the one-move deque containing the group-theoretic inverse of \(a_t\),
and \(\Vert\) denotes deque concatenation (push-front).

\paragraph{Stopping rule and outcomes.}
An episode \emph{solves} if \(d(s_t)=0\) for some \(t\le \texttt{max\_attempts}\);
otherwise it is a \emph{failure}. The number of attempts used is
\[
T \;=\;
\begin{cases}
\min\{t:\ d(s_t)=0\}, & \text{if solved within } \texttt{max\_attempts},\\[2pt]
\infty, & \text{otherwise}.
\end{cases}
\]
For reporting, the solved-attempts histogram is \(H(k)=\bigl|\{\,\text{episodes with }T=k\,\}\bigr|\).
Episode-level proportions such as \(P(1)\) and \(P(\le 3)\) use the unconditional denominator
(all \(N\) episodes; unsolved contribute \(0\)). The \emph{Avg@All} statistic treats failures
as \(\texttt{max\_attempts}\) when averaging attempts across episodes; \emph{Med@Solved} is the
median of \(T\) over solved episodes only (NA if none).

\subsubsection{Generation \& Fairness Controls}
\label{app:ctrl:lcurve}

\paragraph{Deterministic scrambles \& shared decoding.}
Each episode \(i\) is generated with a fixed scramble depth \(d\) and a deterministic seed equal to the episode index (\texttt{random\_seed=idx}). All models use the same prompt, renderer settings, temperature \(=0.0\), and a large \texttt{max\_new\_tokens}, ensuring identical inputs and decoding policy.

\begin{table}[t]
    \centering
    \small
    \begin{tabular}{ccc}
        \toprule
        \textbf{Depth} ($d$) & \textbf{Closed-Loop Horizon} & \textbf{Recovery Budget} \\
        \midrule
        1 & 1 step & 4 \\
        2 & 2 steps & 5 \\
        3 & 3 steps & 6 \\
        4 & 4 steps & 7 \\
        \bottomrule
    \end{tabular}
    \caption{Task parameters by depth. The \textbf{Closed-Loop Horizon} defines the maximum steps allowed in the standard Step-by-Step task (strictly equal to $d$). The \textbf{Recovery Budget} defines the maximum additional moves allowed to solve the cube after the first deviation in the Learning Curve test.}
    \label{tab:horizons_attempts}
\end{table}

\paragraph{Canonical option set, with a minimal fallback.}
From state \(s_t\), we compute the \emph{progress set}
\[
G(s_t)=\{\,m:\ d(m(s_t))<d(s_t)\ \text{ or }m(s_t)\text{ is solved}\,\},
\]
using the same oracle distance \(d(\cdot)\) as for scoring. When feasible, we build a \emph{canonical} MCQ with exactly one progress-making option sampled from \(G(s_t)\) and three distinct distractors sampled from \(\mathcal{M}\setminus G(s_t)\) (never equal to the correct move). If \(|\mathcal{M}\setminus G(s_t)|<3\) (rare), we use a \emph{fallback} that samples the remainder from \(\mathcal{M}\setminus\{\text{correct}\}\) without enforcing progress/non-progress; options are still unique and evaluable.

\paragraph{Uniform letter shuffling, independent RNG.}
Options are uniformly permuted into A-D using a full-entropy RNG (System Random) that is independent of both the scramble seed and episode index. This prevents fixed letter positions from correlating with correctness and reduces positional cueing. We do not otherwise regularize or audit slot frequencies; independence plus sample size keeps slot priors approximately flat in expectation.

\paragraph{Parse protocol \& penalties.}
The model must output exactly one letter via either \verb|<ANSWER>X</ANSWER>| or \verb|ANSWER: X| (case-insensitive), \(X\in\{\mathrm{A},\mathrm{B},\mathrm{C},\mathrm{D}\}\). Any other output is a \emph{parse-fail}. Parse-fails count as attempts and leave the environment unchanged, applying a consistent penalty without introducing spurious transitions.

\paragraph{Fixed control policy (accept-progress).}
The same policy is applied to every model’s choice \(a_t\): (i) if \(a_t\) matches the teacher head, apply and advance the plan; (ii) else if \(a_t\in G(s_t)\), apply and replan from the new state; (iii) else apply and prepend \(a_t^{-1}\) to the plan. This creates a uniform opportunity to recover from near-misses and a uniform consequence for regressions.

\paragraph{Attempts budget parity \& stopping.}
All models share the same \texttt{max\_attempts} and stopping rule (solve iff \(d(s)=0\) within budget). Aggregate metrics (SR, \(P(1)\), \(P(\le3)\), Med@Solved, Avg@All) are computed over the same episode set with the unconditional denominator, so unsolved runs contribute \(0\) to proportion metrics and \(\texttt{max\_attempts}\) to Avg@All.

\paragraph{Scope of randomness.}
We intentionally resample distractors each attempt (subject to the uniqueness and canonical/fallback rules) rather than fixing distractor identities across episodes or models. This avoids overfitting to specific move tuples while keeping evaluation comparable through shared scrambles, shared decoding, and a fixed control policy.

\clearpage
\section{Detailed Prompts}
\label{app:prompts}
\subsection{Cube Face Reconstruction}
\label{app:prompts:reconstruction}

\begin{systemprompt}
    You are a precise AI assistant with advanced image analysis capabilities.
    
    The official Rubik's Cube sticker colors and their RGB values are:
    \begin{itemize}
        \item White: (255,255,255)
        \item Yellow: (255,255,0)
        \item Red: (255,0,0)
        \item Orange: (255,128,0)
        \item Blue: (0,0,255)
        \item Green: (0,255,0)
    \end{itemize}
    
    When given the cube-net image:
    \begin{enumerate}
        \item Locate the 'Front' face (center of the net).
        \item Extract its 3*3 sticker colors, row by row.
        \item Verify each cell by cross-referencing its RGB against the table above.
        \item Output strictly in this format:\\
            ANSWER:\\
            Row 1: [Color1, Color2, Color3]\\
            Row 2: [Color4, Color5, Color6]\\
            Row 3: [Color7, Color8, Color9]\\
        \item Finally, append exactly this line:\\
            Answer verified for correctness.
    \end{enumerate}

    Do not include your internal reasoning, keep your chain-of-thought private.
\end{systemprompt}

\begin{userprompt}
    Here is the image of the Rubik's Cube laid out in a net diagram. The front face is labeled as 'Front'. Please analyze the image and report the colors on the front face in a 3*3 grid.
\end{userprompt}

\subsection{Cross-Modal Verification}
\label{app:prompts:verification}

\begin{systemprompt}
You are a specialized Rubik's Cube verification assistant with independent textual and visual analysis capabilities.
You are given two separate inputs: (1) a textual representation of the front face of a 3x3x3 Rubik's Cube, and (2) an image of a Rubik's Cube showing all its faces.

\textbf{Tasks:}
\begin{enumerate}
  \item Independently identify the front face in the image.
  \item Compare the nine color positions on the front face from the image with the provided textual representation. Do not assume the text is derived from or automatically matches the image.
  \item If all nine positions match exactly in color, respond with \texttt{Answer: Yes}. If any position does not match, respond with \texttt{Answer: No}.
\end{enumerate}

\textbf{Important:} Always perform the verification independently; do not default to \texttt{Answer: Yes} without a thorough comparison.
\end{systemprompt}

\begin{userprompt}
\textbf{Inputs}
\begin{enumerate}
  \item A textual representation of the front face of a 3x3x3 Rubik's Cube: \verb|{front_face}|
  \item An image of a Rubik's Cube showing all its faces.
\end{enumerate}

\textbf{Task}
\begin{itemize}
  \item Determine the front face of the Rubik's Cube in the image.
  \item Compare the colors of the front face with the provided textual representation.
  \item Output \texttt{Answer: Yes} if they match exactly in all nine positions, or \texttt{Answer: No} if there is any mismatch.
\end{itemize}
\end{userprompt}

\subsection{Optimal Move Prediction}
\label{app:prompts:move_selection}
\subsubsection{Image + Text Prompt}
\label{app:prompts:move_selection:image_text}

\begin{systemprompt}
    You are an expert Rubik's Cube solver. You will be given a \textbf{TEXTUAL} cube state, an \textbf{IMAGE} of the cube, and four candidate moves (A–D). The cube is exactly 1 move from solved. Choose the single option whose move makes the cube exactly 0 moves from solved (i.e., solves it). \\
    
    \textbf{CONTRACT (read carefully):}
    \begin{itemize}
        \item Exactly ONE of A/B/C/D is correct. Never answer AMBIGUOUS.
        \item Output must be \textbf{EXACTLY} one of A, B, C, or D wrapped in \texttt{<ANSWER>} tags. Nothing else.
        \item Your entire output MUST match this regex: \verb|^\\s*<ANSWER>\\s*[ABCD]\\s*</ANSWER>\\s*$|
        \item Do NOT include explanations, move strings (e.g., R, U2, R'), or any extra text inside or outside the tags.
    \end{itemize}
    
    \textbf{Notation reminder:}
    \begin{itemize}
        \item X  = 90° clockwise quarter-turn of face X.
        \item X' = 90° counter-clockwise quarter-turn of face X.
        \item X2 = a \emph{single} 180° half-turn of face X (not two 90s).
    \end{itemize}
    
    \textbf{Grounding \& authority (internal):}
    \begin{itemize}
        \item Faces are identified by their centers (isomorphic color scheme).
        \item If TEXT and IMAGE disagree, the \textbf{TEXTUAL} state is authoritative; the IMAGE is for reference only.
    \end{itemize}
    
    \textbf{Parsing rules (internal; do not restate):}
    \begin{itemize}
        \item If the textual state is an ASCII net with tokens [w,y,o,r,g,b], map tokens via centers: w$\to$white, y$\to$yellow, o$\to$orange, r$\to$red, g$\to$green, b$\to$blue. Do NOT invent or output Kociemba strings.
        \item If the textual state is already URFDLB facelets, use it directly.
    \end{itemize}
    
    \textbf{Decision procedure (internal only):}
    \begin{enumerate}
        \item Interpret the state from \textbf{TEXT} (authoritative) and consult the \textbf{IMAGE} if helpful.
        \item Mentally apply each candidate (A–D) to this state.
        \item Select the single candidate that yields the solved cube (distance = 0).
        \item If a tie somehow occurs, break ties by letter: $A \prec B \prec C \prec D$.
        \item Think silently; do not reveal reasoning.
        \item Output only the LETTER in the required format.
    \end{enumerate}
\end{systemprompt}

\begin{userprompt}
    Textual Cube State (authoritative): \\
    \verb|{textual_representation}|
    
    Candidate moves: \\
    A: \verb|{move_A}| \\
    B: \verb|{move_B}| \\
    C: \verb|{move_C}| \\
    D: \verb|{move_D}| \\
    Reply with exactly one line: \\
    \texttt{<ANSWER> A/B/C/D </ANSWER>}
\end{userprompt}

\subsubsection{Text Only Prompt}
\label{app:prompts:move_selection:text}
\begin{systemprompt}
    You are an expert Rubik's Cube solver. You will be given a \textbf{TEXTUAL} cube state and four candidate moves (A–D). The cube is exactly 1 move from solved. Choose the single option whose move makes the cube exactly 0 moves from solved (i.e., solves it). \\
    
    \textbf{CONTRACT (read carefully):}
    \begin{itemize}
        \item Exactly ONE of A/B/C/D is correct. Never answer AMBIGUOUS.
        \item Output must be \textbf{EXACTLY} one of A, B, C, or D wrapped in \texttt{<ANSWER>} tags. Nothing else.
        \item Your entire output MUST match this regex: \verb|^\\s*<ANSWER>\\s*[ABCD]\\s*</ANSWER>\\s*$|
        \item Do NOT include explanations, intermediate moves, or any other text.
    \end{itemize}
    
    \textbf{Notation reminder:}
    \begin{itemize}
        \item X  = 90° clockwise quarter-turn of face X.
        \item X' = 90° counter-clockwise quarter-turn of face X.
        \item X2 = a \emph{single} 180° half-turn of face X (not two 90s).
    \end{itemize}
    
    \textbf{Grounding \& authority (internal):}
    \begin{itemize}
        \item The \textbf{TEXTUAL} state is the single source of truth. Faces are identified by their centers.
    \end{itemize}
    
    \textbf{Parsing rules (internal; do not restate):}
    \begin{itemize}
        \item If the textual state is an ASCII net with tokens [w,y,o,r,g,b], map tokens via centers: w$\to$white, y$\to$yellow, o$\to$orange, r$\to$red, g$\to$green, b$\to$blue. Do NOT invent or output Kociemba strings.
        \item If the textual state is already URFDLB facelets, use it directly.
    \end{itemize}
    
    \textbf{Decision procedure (internal only):}
    \begin{enumerate}
        \item Interpret the state from TEXT using the center mapping.
        \item Mentally apply each candidate (A–D) to this state.
        \item Select the single candidate that yields the solved cube (distance = 0).
        \item If a tie somehow occurs, break ties by letter: $A \prec B \prec C \prec D$.
        \item Think silently; do not reveal reasoning.
        \item Output only the LETTER in the required format.
    \end{enumerate}
\end{systemprompt}

\begin{userprompt}
    Textual Cube State (ground truth): \\
    \verb|{textual_representation}| \\
    Candidate moves: \\
    A: \verb|{move_A}| \\
    B: \verb|{move_B}| \\
    C: \verb|{move_C}| \\
    D: \verb|{move_D}| \\
    Reply with exactly one line: \\
    \texttt{<ANSWER> A/B/C/D </ANSWER>}
\end{userprompt}

\subsubsection{Image Only Prompt}
\label{app:prompts:move_selection:image}

\begin{systemprompt}
    You are an expert Rubik's Cube solver. You will be given an \textbf{IMAGE} of the cube and four candidate moves (A–D). The cube is exactly 1 move from solved. Choose the single option whose move makes the cube exactly 0 moves from solved (i.e., solves it). \\
    
    \textbf{CONTRACT (read carefully):}
    \begin{itemize}
        \item Exactly ONE of A/B/C/D is correct. Never answer AMBIGUOUS.
        \item Output must be \textbf{EXACTLY} one of A, B, C, or D wrapped in \texttt{<ANSWER>} tags. Nothing else.
        \item Your entire output MUST match this regex: \verb|^\\s*<ANSWER>\\s*[ABCD]\\s*</ANSWER>\\s*$|
        \item Do NOT include explanations, move strings, or any other text.
    \end{itemize}
    
    \textbf{Notation reminder:}
    \begin{itemize}
        \item X  = 90° clockwise quarter-turn of face X.
        \item X' = 90° counter-clockwise quarter-turn of face X.
        \item X2 = a \emph{single} 180° half-turn of face X (not two 90s).
    \end{itemize}
    
    \textbf{Grounding \& authority (internal):}
    \begin{itemize}
        \item No textual state is provided. Infer the state from the \textbf{IMAGE} only. Faces are identified by their centers; the cube is solved when every face matches its center color.
    \end{itemize}
    
    \textbf{Parsing rules (internal; do not restate):}
    \begin{itemize}
        \item None (no textual state). Use the pixels only.
    \end{itemize}
    
    \textbf{Decision procedure (internal only):}
    \begin{enumerate}
        \item Infer the current state from the IMAGE using center colors.
        \item Mentally apply each candidate (A–D) to this state.
        \item Select the single candidate that yields the solved cube (distance = 0).
        \item If a tie somehow occurs, break ties by letter: $A \prec B \prec C \prec D$.
        \item Think silently; do not reveal reasoning.
        \item Output only the LETTER in the required format.
    \end{enumerate}
\end{systemprompt}

\begin{userprompt}
    Candidate moves: \\
    A: \verb|{move_A}| \\
    B: \verb|{move_B}| \\
    C: \verb|{move_C}| \\
    D: \verb|{move_D}| \\
    Reply with exactly one line: \\
    \texttt{<ANSWER> A/B/C/D </ANSWER>}
\end{userprompt}

\subsection{Reflections}
\label{app:prompts:reflections}
This section explains how we run \emph{reflection-guided re-answering} on the static move-selection task. Each reflection trial begins from the same \textbf{Image+Text} prompt we use for move prediction test (Prompt~\ref{app:prompts:move_selection:image_text}). Concretely, every item shows (i) a rendered image of the cube for context and (ii) the authoritative textual cube state, along with four candidate move.

\subsubsection{Guided Unredacted Reflection}
\label{app:prompts:reflection:unredacted}
\begin{systemprompt}
    You are an expert Rubik's Cube solver. You previously answered the following multiple-choice question incorrectly.Your task is to reflect on your mistake, understand why it occurred, and improve your reasoning for the future.
    
    Complete the following steps clearly and concisely:
    \begin{enumerate}
        \item Explain why you answered incorrectly.
        \item List keywords describing your mistake type from most general to most specific.
        \item Solve the problem again step-by-step using the correct answer provided.
        \item Write detailed instructions to avoid making this mistake again (No more than 200 words).
        Provide general advice to improve your (Large Language Model) performance on similar problems in the future.
    \end{enumerate}
\end{systemprompt}

\begin{userprompt}
    The cube's textual state was: \verb|{cube_state}|. An image of the cube in this state is attached. The cube was only one move away from being solved.
    
    The four options were: \\
    A: \verb|{option_A}| \\
    B: \verb|{option_B}| \\
    C: \verb|{option_C}| \\
    D: \verb|{option_D}| \\
    
    You chose the option: \verb|{model_choice}|, which did not solve the cube. The correct solution was option: \verb|{correct_answer}|.
\end{userprompt}

\subsubsection{Guided Redacted Reflection}
\label{app:prompts:reflection:redacted}
\begin{systemprompt}
    You are an expert Rubik's Cube solver. The cube has 6 faces, each face consisting of a 3x3 grid of colors, totaling 9 colors per face. The faces are labeled as follows: Front (F), Back (B), Left (L), Right (R), Up (U), and Down (D). On each face, colors are arranged in 3 rows, each containing 3 colors. Colors are read from left to right within each row, starting from the top row and proceeding downward. Thus, each face has exactly 9 colors in a fixed order.
    
    You previously answered the following multiple-choice question incorrectly.
    
    Your task is to reflect on your mistake, understand why it occurred, and improve your reasoning for the future.
    
    Complete the following steps clearly and concisely:
    \begin{enumerate}
        \item Explain why you answered incorrectly.
        \item List keywords describing your mistake type from most general to most specific.
        \item Solve the problem again step-by-step using the correct answer provided.
        \item Write detailed instructions to avoid making this mistake again (No more than 200 words).
        \item Provide general advice to improve your (Large Language Model) performance on similar problems in the future.
    \end{enumerate}
\end{systemprompt}

\begin{userprompt}
    The cube's textual state was: \verb|{cube_state}|. An image of the cube in this state is attached.
    
    The cube was only one move away from being solved.
    
    The four options were: \\
    A: \verb|{option_A}| \\
    B: \verb|{option_B}| \\
    C: \verb|{option_C}| \\
    D: \verb|{option_D}| \\
    
    You chose the option: \verb|{model_choice}|, which did not solve the cube. The correct solution was option: [REDACTED]. Correct answer is hidden, to avoid answer leakage into the reflection.
\end{userprompt}

\subsection{Optimal-Move Adherence vs Depth}
\label{app:prompts:stepbystep}

\begin{systemprompt}
You are an expert Rubik's Cube solver. Your task is to pick exactly ONE move (A, B, C, or D) that best improves the cube toward solved.

\textbf{Context:}
- The cube is currently \verb|{n_moves}| moves from solved under God's-distance.
- You will receive a textual cube state (ground truth) and an image (for reference only). Ignore any discrepancies: the text is authoritative.

\textbf{Decision Rule (no ambiguity allowed):}
\begin{enumerate}
  \item For each candidate (A,B,C,D), internally simulate applying the move to the textual state and estimate the resulting God's-distance \(d_1\).
  \item Prefer the move with the \textbf{lowest} \(d_1\) (equivalently, the largest decrease from the current distance).
  \item If multiple candidates yield the same minimal \(d_1\), break ties \textbf{deterministically by letter}: \(A \prec B \prec C \prec D\).
  \item Never output anything other than A/B/C/D.
\end{enumerate}

\textbf{Input Rules:}
- Use the textual state exclusively for simulation and evaluation; do not use the image to break ties.
- Think step-by-step internally but do \textbf{not} reveal your chain-of-thought.

\textbf{Output Format (STRICT):}
Return exactly one line containing only this XML tag:

\verb|<ANSWER> A/B/C/D </ANSWER>|

No explanations, no extra text, no AMBIGUOUS.
\end{systemprompt}

\begin{userprompt}
The cube is \verb|{n_moves}| moves away from being solved.

\textbf{Textual Cube State (Ground Truth):}
\verb|{textual_representation}|

\textbf{Candidate Moves:}
A: \verb|{move_A}|\\
B: \verb|{move_B}|\\
C: \verb|{move_C}|\\
D: \verb|{move_D}|

\textbf{Output exactly one line:}\\
\verb|<ANSWER> X </ANSWER>|
\end{userprompt}

\subsection{Causal Move-Effect Probe}
\label{app:prompts:move_effect}

\begin{systemprompt}
You are an expert Rubik's Cube evaluator. For \textbf{each} option (A--D), predict how applying that move to the textual cube state changes God's-distance.

\textbf{Rules}
\begin{itemize}
  \item Textual cube state is ground truth; ignore any images.
  \item Use face centers to map colors to faces.
  \item Parse moves in Singmaster notation: quarter \(X\), inverse \(X'\), half \(X2\); apply sequences left-to-right.
  \item Report only the \emph{sign}: \texttt{DECREASE}, \texttt{NO\_CHANGE}, or \texttt{INCREASE}.
  \item If a move is invalid or cannot be parsed, use \texttt{NO\_CHANGE}.
  \item Think step-by-step internally but do not reveal reasoning.
\end{itemize}

\textbf{Output (STRICT):} exactly four lines, in order A,B,C,D:
\begin{verbatim}
    |<A> DECREASE|NO_CHANGE|INCREASE </A>|
    |<B> DECREASE|NO_CHANGE|INCREASE </B>|
    |<C> DECREASE|NO_CHANGE|INCREASE </C>|
    |<D> DECREASE|NO_CHANGE|INCREASE </D>|
\end{verbatim}
\end{systemprompt}

\begin{userprompt}
\textbf{Face Centers}\\
U: \verb|{U_color}| \quad R: \verb|{R_color}| \quad F: \verb|{F_color}|\\
D: \verb|{D_color}| \quad L: \verb|{L_color}| \quad B: \verb|{B_color}|

\textbf{Textual Cube State (Ground Truth)}\\
\verb|{state_text}|

\textbf{Candidate Moves}\\
A: \verb|{move_A}| \\
B: \verb|{move_B}| \\
C: \verb|{move_C}| \\
D: \verb|{move_D}|

\textbf{Return exactly four lines (uppercase labels), in this order:}\\
\begin{verbatim}
    |<A> DECREASE|NO_CHANGE|INCREASE </A>|
    |<B> DECREASE|NO_CHANGE|INCREASE </B>|
    |<C> DECREASE|NO_CHANGE|INCREASE </C>|
    |<D> DECREASE|NO_CHANGE|INCREASE </D>|
\end{verbatim}
\end{userprompt}

\subsection{Recovery dynamics}
\label{app:prompts:recovery}

\begin{systemprompt}
You are an expert Rubik's Cube solver. Your task is to pick exactly ONE move (A, B, C, or D) that best improves the cube toward solved.

\textbf{Context:}
- The cube is currently \verb|{n_moves}| moves from solved under God's-distance.
- You will receive a textual cube state (ground truth) and an image (reference only). If they disagree, the text is authoritative.

\textbf{Decision Rule (no ambiguity):}
\begin{enumerate}
  \item For each candidate (A,B,C,D), internally simulate applying the move to the textual state and estimate the resulting God's-distance \(d_1\).
  \item Choose the move with the \textbf{lowest} \(d_1\) (largest decrease from current distance).
  \item If multiple candidates tie at the minimal \(d_1\), break ties \textbf{deterministically by letter}: \(A < B < C < D\).
  \item Never output anything other than A/B/C/D.
\end{enumerate}

\textbf{Input Rules:}
- Use the textual state exclusively for simulation and evaluation; do not use the image to break ties.
- Think step-by-step internally but do \textbf{not} reveal your chain-of-thought.

\textbf{Output Format (STRICT):}
Return exactly one line containing only this XML tag:

\verb|<ANSWER> A/B/C/D </ANSWER>|

No explanations, no extra text, no ambiguous output.
\end{systemprompt}

\begin{userprompt}
The cube is \verb|{n_moves}| moves away from being solved.

\textbf{Textual Cube State (Ground Truth):}
\verb|{textual_representation}|

\textbf{Candidate Moves:}
A: \verb|{move_A}| \\
B: \verb|{move_B}| \\
C: \verb|{move_C}| \\
D: \verb|{move_D}|

\textbf{Output exactly one line:}\\
\verb|<ANSWER> X </ANSWER>|
\end{userprompt}

\clearpage
\section{Additional Quantitative Results}
\label{app:quantitative_results}

\subsection{Per-Depth Correlation Plots}
\label{app:quantitative_results:correlation}
\begin{figure*}[t]
    \centering
    \begin{subfigure}[b]{0.32\textwidth}
        \centering
        \includegraphics[width=\linewidth]{Images/kappa_vs_ta_d1.pdf}
        \caption{Depth $d=1$}
        \label{fig:kappa_d1}
    \end{subfigure}
    \hfill 
    \begin{subfigure}[b]{0.32\textwidth}
        \centering
        \includegraphics[width=\linewidth]{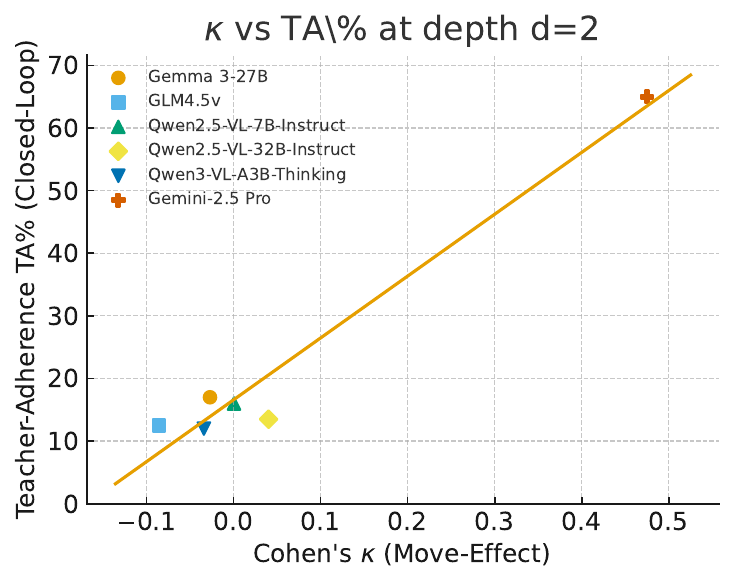}
        \caption{Depth $d=2$}
        \label{fig:kappa_d2}
    \end{subfigure}
    \hfill 
    \begin{subfigure}[b]{0.32\textwidth}
        \centering
        \includegraphics[width=\linewidth]{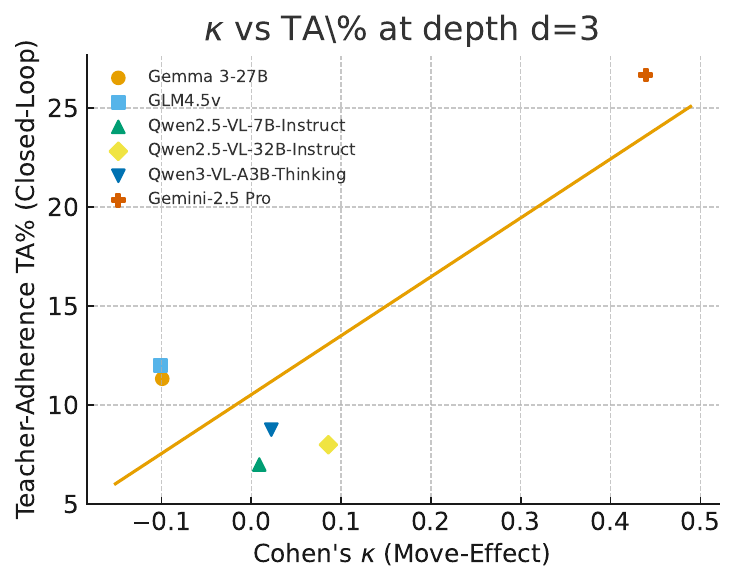}
        \caption{Depth $d=3$}
        \label{fig:kappa_d3}
    \end{subfigure}
    
    \caption{\textbf{Correlation between Pre-action Causal Evaluation ($\kappa$) and Closed-Loop Teacher Adherence (TA\%) across depths.} At $d=1$, the association is strong ($r=0.997$), showing that models capable of causal forecasting perform better at control. However, as depth increases ($d=2, 3$), the relation weakens as open-weight models cluster near chance performance while the frontier model degrades.}
    \label{fig:kappa_vs_ta_all}
\end{figure*}

To investigate the link between local causal understanding and sequential control, we plot the relationship between the \textit{Causal Move-Effect} metric (Chance-corrected Cohen's $\kappa$) and \textit{Closed-Loop Teacher Adherence} (TA\%) across scramble depths $d \in \{1, 2, 3\}$.

As illustrated in Figure~\ref{fig:kappa_vs_ta_all}, the strength of this association varies significantly with task horizon:

\begin{itemize}
    \item \textbf{Depth 1 (Strong Alignment):} At $d=1$, we observe a near-perfect linear correlation ($r=0.997$) between pre-action evaluation and control performance. This indicates that for single-step problems, a model's ability to forecast move consequences ($\kappa > 0$) directly translates to successful execution.
    
    \item \textbf{Depths 2 and 3 (Decoupling):} As the sequence length increases, the linear relationship weakens. At $d=2$ and $d=3$, open-weight models cluster near the origin (indicating chance-level performance on both metrics), while the frontier model (Gemini 2.5 Pro) degrades but remains distinct. This "collapse" suggests that while local causal understanding is a prerequisite for control, it is insufficient to prevent trajectory divergence in multi-step planning without robust error recovery mechanisms.
\end{itemize}

\subsection{Unredacted Reflection Results}
\label{app:quantitative_results:unredacted_reflection}

\begin{table}[t]
    \centering
    \small
    \begin{tabular}{@{}l c c c c c@{}}
        \toprule
        & & \multicolumn{4}{c}{\textbf{Guided (Unredacted)}} \\
        \cmidrule(lr){3-6}
        \textbf{Model} & \textbf{Init. (\%)} & \textbf{Final (\%)} & \textbf{$\Delta$} & \textbf{EFR (\%)} & \textbf{OTR (\%)} \\
        \midrule
        Gemma 3-27B                & 22.00 & 81.00  & +59.00 & 81.70  & 20.70 \\
        GLM-4.5v                   & 31.00 & 100.00 & +69.00 & 100.00 & 0.00 \\
        Llama 4-Scout-17B:16E      & 18.00 & 100.00 & +82.00 & 100.00 & 0.00 \\
        Qwen 2.5-VL-7B-Inst.       & 17.00 & 59.00  & +42.00 & 59.00  & 41.20 \\
        Qwen 2.5-VL-32B-Inst.      & 24.00 & 70.00  & +46.00 & 71.80  & 36.43 \\
        Qwen3-VL-A3B-Thinking      & 23.00 & 72.00  & +50.00 & 71.80  & 27.34 \\
        \bottomrule
    \end{tabular}
    \caption{\textbf{Guided (Unredacted) Reflection Results.} This table presents the "leaky" regime where the reflection prompt includes the correct ground-truth option as a hint. This serves as an upper-bound ablation, showing that models can achieve near-perfect correction (EFR up to 100\%) when given oracle guidance, in contrast to the modest gains seen in the label-safe regime.}
    \label{tab:reflection-unredacted}
\end{table}

In the main text (Section 4.3), we presented "Guided (Redacted)" reflection, where the model must critique its own output without access to the ground truth. Here, we present the results of the "Guided (Unredacted)" ablation. In this regime, the reflection prompt explicitly includes the correct option as a hint, serving as an oracle upper bound to determine if models can recognize and justify the correct move when it is pointed out to them.

Table~\ref{tab:reflection-unredacted} summarizes these findings. In stark contrast to the label-safe setting i.e., where gains were modest ($+6$\% for Llama 4) or negative ($-15$\% for GLM-4.5v) the unredacted setting yields massive improvements across all architectures.

\begin{itemize}
    \item \textbf{Perfect Rationalization:} GLM-4.5v and Llama 4-Scout-17B achieve \textbf{100\% Final Accuracy} (gains of +69.0 and +82.0 points, respectively). This confirms that these models possess sufficient world knowledge to \textit{verify} a solution when given the answer, even though they fail to \textit{derive} it autonomously.
    \item \textbf{Broad Improvement:} Even the models that struggled most in the standard setting show strong recovery. Gemma 3-27B, which showed zero net gain in the redacted setting, jumps from 22\% to 81\% accuracy in the unredacted setting. Similarly, the Qwen family sees improvements ranging from +42 to +50 points.
\end{itemize}

These results highlight the "Post-Hoc Rationalization" capabilities of MLLMs. The discrepancy between the Redacted and Unredacted scores indicates that the bottleneck in spatial reasoning is not the inability to understand the move mechanics, but the lack of a robust internal verification signal. When that signal is provided externally (the oracle label), models can successfully align their reasoning, but without it, they are prone to overthinking and instability.

\end{document}